\documentclass{article}

\AtBeginDocument{%
  }

\usepackage{arxiv}

\usepackage[linesnumbered,ruled,vlined]{algorithm2e}
\usepackage{algorithmic}

\SetCommentSty{mycommfont}
\usepackage{array}
\usepackage{multirow}
\usepackage{amssymb, amsmath}

\usepackage{url}

\usepackage{enumitem}
\usepackage{algorithmic}

\usepackage{mathtools}

\usepackage{textcomp}
\usepackage{xcolor}

\usepackage{booktabs}
\usepackage{comment}
\usepackage{subfig}

\usepackage{distribalgo}
\usepackage{acronym}

\usepackage{multirow}
\usepackage{xcolor}

\newcommand{\rbnote}[1]{ {\textcolor{orange} { ***Rahul: #1 }}}

\newcommand{\krnote}[1]{ {\textcolor{yellow} { ***Jayaram: #1}}}

\newcommand{\nvnote}[1]{ {\textcolor{orange} { ***Nalini: #1 }}}

\newcommand{\changes}[1]{ {\textcolor{blue} {#1}}}

\newcommand{\ours}{FLIPS}

\newcommand{\nip}[1]{\vspace{1ex}\noindent\textbf{#1}}

\acrodef{FL}{Federated Learning}
\acrodef{FFL}{Framework for Federated Learning}
\acrodef{AI}{artificial intelligence}
\acrodef{ML}{machine learning}
\acrodef{MLaaS}{machine learning as a service}
\acrodef{DL}{deep learning}
\acrodef{DNN}{deep neural network}
\acrodef{ConvNet}{convolutional neural network}
\acrodef{SGX}{Software Guard Extensions}
\acrodef{SEV}{Secure Encrypted Virtualization}
\acrodef{EVM}{encrypted virtual machine}
\acrodef{TDX}{Trust Domain Extensions}
\acrodef{TEE}{Trusted Execution Environment}
\acrodef{MEE}{Memory Encryption Engine}
\acrodef{OS}{operating system}
\acrodef{SoC}{System-on-Chip}
\acrodef{SP}{Secure Processor}
\acrodef{PEF}{Protected Execution Facility}
\acrodef{HPVS}{Hyper Protect Virtual Servers}
\acrodef{FedSGD}{Federated Stochastic Gradient Descent}
\acrodef{FedAvg}{Federated Averaging}
\acrodef{IoT}{Internet of Thing}
\acrodef{VM}{virtual machine}
\acrodef{SME}{Secure Memory Encryption}
\acrodef{AES}{Advanced Encryption Standard}
\acrodef{ASID}{Address Space Identifier}
\acrodef{MSE}{Mean Squared Error}
\acrodef{OVMF}{Open Virtual Machine Firmware}
\acrodef{IoT}{Internet of Things}

\acrodef{PDH}{Platform Diffie-Hellman Public Key}
\acrodef{GODH}{Guest Owner Diffie-Hellman Public Key}
\acrodef{CEK}{Chip Endorsement Key}
\acrodef{ASK}{AMD SEV Signing Key}
\acrodef{ARK}{AMD Root Key} 
\acrodef{OCA}{Owner Certificate Authority} 
\acrodef{TIK}{Transport Integrity Key}
\acrodef{TEK}{Transport Encryption Key}
\acrodef{KEK}{Key Encryption Key}
\acrodef{KIK}{Key Integrity Key}
\acrodef{VEK}{VM Encryption Key}

\acrodef{DHKE}{Diffie-Hellman Key Exchange}
\acrodef{HE}{Homomorphic Encryption}
\acrodef{SMC}{Secure Multi-Party Computation}
\acrodef{DP}{Differential Privacy}
\acrodef{CDP}{Centralized Differential Privacy}
\acrodef{LDP}{Local Differential Privacy}
\acrodef{DDP}{Distributed Differential Privacy}

\begin{document}


\title{FLIPS: Federated Learning using Intelligent Participant Selection}

\author{Rahul Atul Bhope \\
	UC Irvine\\
	\texttt{rbhope@uci.edu} \\
\And
 K. R. Jayaram \\
IBM Research AI\\	
\texttt{jayaramkr@us.ibm.com} \\
\And
Nalini Venkatasubramanian \\
UC Irvine\\	
\texttt{nalini@uci.edu} \\
\And
Ashish Verma \\
Amazon Inc.\thanks{Ashish Verma was affiliated with IBM Research during this work.}\\	
\texttt{draverma@amazon.com} \\
\And
Gegi Thomas \\
IBM Research AI\\	
\texttt{gegi@us.ibm.com} \\
}

\maketitle

\begin{abstract}

This paper presents the design and implementation of \ours, a middleware system to manage
data and participant heterogeneity in federated learning (FL) training workloads. 
In particular, we examine the benefits of label distribution clustering on 
participant selection in federated learning.
\ours\ clusters parties involved in an FL training job 
based on the label distribution of their data apriori, and during FL training, 
ensures that each cluster is equitably represented in the participants selected.
\ours\ can support the most common FL algorithms, including FedAvg, FedProx, FedDyn, FedOpt and 
FedYogi.
To manage platform heterogeneity and dynamic resource availability, 
\ours\ incorporates  a straggler management mechanism to handle changing 
capacities in distributed, smart community applications.
Privacy of label distributions, clustering and participant selection is ensured 
through a trusted execution environment (TEE). 
Our comprehensive empirical evaluation compares \ours\ with random participant selection, as well as 
three other ``smart'' selection mechanisms -- Oort~\cite{Oort-osdi21}, TiFL~\cite{tifl}
and gradient clustering~\cite{pmlr-v139-fraboni21a} using two real-world datasets, two benchmark datasets, two different non-IID 
distributions and three common FL algorithms (FedYogi, FedProx and FedAvg). 
We demonstrate that \ours\ significantly improves convergence, achieving higher
accuracy by 17-20 percentage points with 20-60\% lower communication costs, and these benefits endure
in the presence of straggler participants.


\end{abstract}

\section{Introduction}~\label{sec:intro}

\noindent{ Federated Learning (FL)~\cite{kairouz2019advances} is the process by which multiple participants (parties)
collaborate to train a common machine learning (ML) model, without sharing data 
among themselves or with a centralized cloud-hosted machine learning service.
FL allows parties to retain private data within their controlled domains. 
Only model updates are \emph{typically} shared to a central aggregation server hosted by one of
the parties or a cloud service provider. This, along with transformations applied to model
updates (e.g., homomorphic encryption~\cite{jayaram-cloud2020} and 
addition of noise for differential privacy~\cite{abadi-diffpriv}) make FL \emph{privacy preserving}.

\noindent{\bf Why FL?} From a participant perspective, a key goal of FL is to \emph{access diverse training data} to enhance
the robustness of machine learning models by promoting generalization, outlier detection and 
bias mitigation. FL allows the training of machine learning models using data distributed across multiple devices or 
edge nodes from different demographies and geographical locations. This distributed nature allows for 
diverse datasets, as each device may have unique data reflecting various user behaviors, preferences, or contexts. 
The privacy-preserving aspect encourages a wider range of participants to contribute their data, 
including those who may be concerned about sharing personal information or those restricted by 
regulations like HIPAA and GDPR, further leading to a more representative and 
inclusive dataset. 

\noindent{\bf Benefits of Diverse Datasets:} A diverse training dataset provides a broader representation of the real-world scenarios and 
variations that the model may encounter during inference. By exposing the model to a wide range of data samples, 
including different classes, variations, edge cases and outliers, the model can learn more generalized patterns 
and make better predictions on unseen data. 
Outliers can be valuable in identifying 
rare events or unusual patterns that may not be present in a homogeneous dataset.
Also, including diverse samples 
that represent different demographics, backgrounds, and perspectives, models can learn to be 
more equitable and avoid perpetuating bias. It enables the model to make predictions that 
are more representative and fair for a broader range of individuals. 

\noindent{\bf non-IID Data in FL.} The diversity of real-world
entities and their private data also implies that FL techniques and algorithms have to be 
designed to handle non-IID (non Independent and identically distributed) data.
Parties not only have different data items, but also often have wildly different types of data items, 
corresponding to different labels. 
Several leading FL researchers~\cite{fieldguide,kairouz2019advances} have noted that the 
presence of IID data is the exception rather than the norm~\cite{fieldguide}.

\noindent{\bf Intermittent Participation.} FL, in general, is characterized by \emph{intermittent} participation, which
means that for every FL round, each party trains at its convenience, 
or feasibility. This may be when devices are connected to power in the case of mobile phones, tablets and laptops
(FL over edge devices); when (local) resource utilization from other computations is low and when there are no pending jobs with higher priority (both in edge and datacenter use cases). 
The aggregator expects to hear from the parties \emph{eventually} - typically over several 
minutes or hours. 
Parties in FL jobs are often highly unreliable and are expected to drop and rejoin frequently. 

\noindent{\bf Participant Selection:} Due to the intermittent nature of parties, existing FL algorithms 
like FedAvg, FedProx, FedMA, FedDyn, etc. only select a subset of parties in each round, 
often employing randomization to select parties~\cite{fieldguide,kairouz2019advances}. 
Random selection eventually offers each party an 
opportunity to participate in the FL job, but does not take into account the type
of data present at each party, and does not ensure that parties with diverse
datasets are selected in \emph{each round}. There is also significant
empirical evidence that Non-IID data combined with random selection significantly increases the time taken for models
to converge~\cite{noniid-fl,super-dirichlet,bingsheng-noniid} 
(we reproduce some of these results in Section~\ref{sec:eval}).

There is increasing recognition that random selection is suboptimal 
for other reasons as well. The selected participant could have compute or communication constraints 
and might actually not be able to participate in the round. There is existing 
research on selecting participants for each round based on the ability to participate, 
amount of data present at each participant, history of reliable participation, 
and communication constraints~\cite{9355774,Oort-osdi21,https://doi.org/10.48550/arxiv.2211.05739}. 
But said research has not considered the label distribution 
among participants. This is unfortunate because it is vital for model generalization in 
FL to equitably consider all participants, including outliers.  
This paper makes the following contributions: 

\begin{itemize}[topsep=0pt,partopsep=0pt,parsep=0pt,itemsep=0pt]
\item \ours, a middleware for effective management of data and platform 
heterogeneity in FL using clustering techniques based on label distributions.
\item An algorithm to use the generated clusters to select diverse participants 
at each round of an FL training job, ensuring that parties are equitably 
represented while offering each party a fair opportunity to participate.
\item A private mechanism using trusted execution environments (TEEs) to cluster participants in FL jobs based on the label distributions of their data and identify the diversity in their data in a secure manner.
\item A thorough empirical evaluation comparing \ours\ with the predominant 
random participant selection, as well as   
two other ``smart'' selection mechanisms -- Oort~\cite{Oort-osdi21}
and gradient clustering~\cite{pmlr-v139-fraboni21a} using two real world datasets, two benchmark datasets, two different non-IID 
distributions and three common FL algorithms (FedYogi, FedProx and FedAvg). 
We demonstrate that \ours\ significantly improves convergence, achieving superior
accuracy with much lower communication costs, and these benefits endure
in the presence of stragglers.
\end{itemize}



\section{Federated Learning and Heterogeneity}~\label{section:benefits}

A typical FL setting consists of parties with local data and an aggregator 
server (when the number of parties is small) or service (e.g., a microservice using Apache Spark
to aggregate model updates when number of parties is large) to orchestrate FL. 
An FL job proceeds over several rounds, also called \emph{synchronization rounds}.
An aggregator typically coordinates the entire FL job, in addition to aggregating 
model updates and distributing the updated model back to the parties.
Co-ordination, includes, agreeing on the following FL job parameters before the job starts:
(i) model architecture (ResNet, EfficientNet, etc), 
(ii) FL algorithm (FedAvg, FedYogi, etc.) and any algorithm specific parameters like
minimum number of parties required for each round,
(iii) how to initialize the global model 
(whether random, or from existing pre-trained models),
(iv) hyperparameters to be 
used (batch size, learning rate etc.), and 
(v) termination criteria, whether the FL job ends after a specific number
of rounds, or when a majority of parties decide that the model is
satisfactory (e.g., has reached a desired level of accuracy).

\begin{figure}[htb]
    \centering
    \includegraphics[width=0.5\columnwidth]{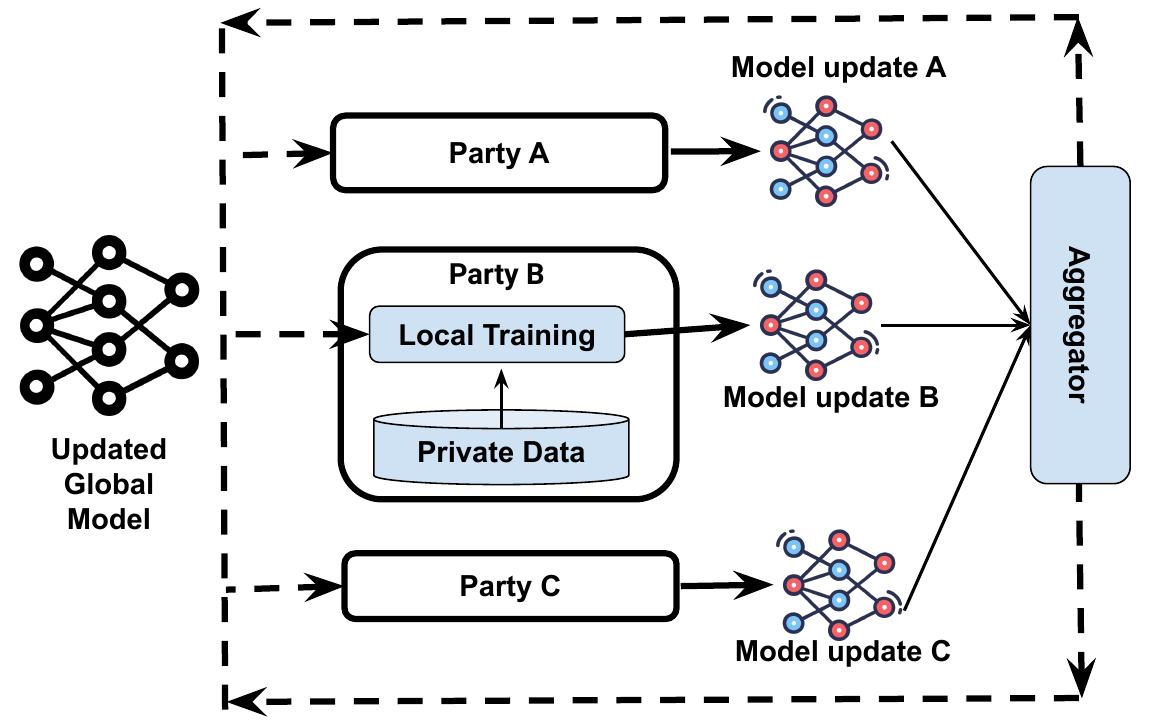}
    \caption{Overview of Federated Learning}
    \label{fig:floverview}
\end{figure}

At each FL job round, each participant trains a local model using its local data. 
The local model is initialized with the global model parameters 
received from the aggregator. The local training process typically involves 
several iterations or epochs (this is part of the hyperparameters
agreed at the start) to improve the model's performance on the local dataset.
After local training, the participants generate model updates,
which typically consist of the updated model parameters or gradients. 
These updates capture the local knowledge learned from the device's data.
The aggregator combines these updates, and applies the aggregated update to 
construct the global model using the optimizer. The new global model 
parameters are sent to the participants selected for the next round.
The FL process typically involves multiple rounds of 
local training, model updates, aggregation, and distribution. 
This iterative process allows the model to be refined and improved over 
time by leveraging the collective intelligence of the participating devices.


\subsection{Common FL Algorithms}

FL algorithms~\cite{fieldguide} primarily differ in how model 
updates are aggregated and the mathematical formula (called the \textsc{optimizer} in machine learning
literature) 
used to apply the aggregated model update to the global model.

For FedAvg~\cite{https://doi.org/10.48550/arxiv.1602.05629}, 
the aggregator selects a random subset $\mathcal{S}^{(r)} \subset \mathcal{S}$ of parties 
for every round $r$. The aggregator in FedAvg
computes the weighted average of all participant updates (gradients):  
$\frac{1}{N} {\sum_{i \in  \mathcal{S}^{(r)}} n_i x_{i}^{(r)}}$ to 
compute the global model update $x^{(r)}$ and update the global model (for the next round)
$m^{(r+1)}$ using SGD as the \textsc{optimizer}. 
This process proceeds for a set number $R$ of rounds or (less typically)
until a majority of the parties vote to terminate. The term $n_i$ in the weighted average is the number of training 
samples at party $i$, and $N$ is the total number of training samples involved in the round, i.e., $N = \sum_{i \in  \mathcal{S}^{(r)}} n_i$. FedAvg is the first widely-deployed FL algorithm but does not result in an optimized global model when the data distribution is not IID.

When the entire set of parties $\mathcal{S}$ is used in every round,
and SGD is the \textsc{optimizer}, we get FedSGD. FedAdam and FedAdagrad are the same FedAvg 
with Adam~\cite{adam} and AdaGrad~\cite{adagrad} as the \textsc{optimizer}, respectively.
FedProx~\cite{fedprox} is a variation of the FedAvg that aims to produce better global models in the presence of non-IID data. 
FedProx includes a Proximal term in the $\textsc{optimizer}$ with penalty parameter $\mu$. 
If $F_k(x_{r,k})$ is the local loss function at a party $k$ at round $r$. Then the local loss function in FedProx includes $\mu$ which then translates to $F_k(x^{(r,k)}) + \frac{\mu}{2} || m^{(r)} - x^{(r,k)} ||$. This brings the model $x^{(r,k)}$ closer to $m^{(r)}$ at each party $k$. By requiring the updates to be close to the starting point, a big $\mu$ could potentially hinder convergence, but a small $\mu$ might have no effect.   

FedYoGi~\cite{Oort-osdi21, reddi-adaptopt}, which has been shown to outperform FedAvg and FedProx~\cite{Oort-osdi21} with non-IID data, uses an adaptive optimizer to update the global model. The server $\textsc{optimizer}$ maintains a per-parameter learning rate, updated based on the history of gradients $gr^{(r,k)} = x^{(r,k)} - m^{(r)}$. This allows the server $\textsc{optimizer}$ to adapt to the local data distributions of the clients. FedYoGi introduces a moving average of gradients term $m_{t} = \beta_{1} * m_{t} + (1 - \beta_{1})*gr^{(r,k)}$ and moving average of squared gradients $v_{t} = \beta_{2}*v_{t} + (1 - \beta_{2})*(gr^{(r,k)})^{2}$, with 2 momentum hyperparameters $\beta_{1}$ and $\beta_{2}$. $m^{(r)}$ is updated by $m^{(r)} - lr * \frac{m_t}{\sqrt{v_t} + eps}$, where $lr$ is the learning-rate and $eps$ is a small constant to prevent division by 0.  

\subsection{Data Heterogeneity : Dealing with Non-IID Data}~\label{sec:randomselectionproblems}


Despite advances in FL, the ability of current techniques to deal with variabilities and 
diversity in real-world data is still limited.
The term non-IID (non-independent and identically distributed) data distributions in the 
context of FL refers to the situation in which the data on each device or node 
taking part in the FL process is not independently and identically distributed 
across all devices. This can happen when data is collected from several 
sources or under various circumstances, resulting in variances in 
the data and/or label distribution on each device. 



In FedAvg, FedProx and FedYoGi at each round $r$, $S^{(r)}$ parties 
are sampled randomly from the given pool. $|S^{(r)}|$ is typically small when compared to $|\mathcal{S}|$,
typically less than 20\% in real deployments~\cite{bonawitz2019towards, huba2022papaya}.
This will result in some rounds, where parties
with similar data distributions are chosen and thus lead to class imbalance, when certain classes 
of data are underrepresented on certain parties. This leads to model divergence in rounds when 
parties with diverse data are chosen, and the model to be biased towards the overrepresented classes. 
To understand why
this happens, it is helpful to consider the centralized learning setting, where the training typically makes
a pass over the entire dataset in every training round (epoch), and outlier and diverse data is considered 
in every training epoch. With random selection, outlier data may get omitted continuously for several rounds
in FL, especially with typical values of $|S^{(r)}|$. 

To further illustrate the significance, we consider a real-world use-case from Senior Care focusing on smartspaces and assisted living (name hidden for anonymity). One application investigated for FL is Arrhythmia detection 
using ECG signals~\cite{932724} from wearables where data exhibits non-IIDness, as more data points are 
recorded for normal heartbeats. Abnormal heartbeats are recorded in devices worn by 
people with heart ailments, a small fraction of all the parties in FL~\cite{932724}. 
With random selection in any FL algorithm, 
there is always a higher chance of selecting a party with majority Normal beats at any round, biasing the model
towards classifying most heartbeats as Normal.

\subsection{Platform Heterogeneity : Stragglers}

 In federated learning (FL) deployments in the real world, Platform Heterogeneity plays a significant role in the convergence of the global model. Platform heterogeneity across different parties causes some of them to be stragglers which exhibit intermittent failure. Stragglers are devices that take longer than expected in an FL environment to complete their local training. These gadgets have the potential to stall the FL process and perhaps fail. Stragglers can appear in real-world FL deployments for a variety of reasons. Among the most popular explanations are:

\begin{itemize}[topsep=0pt,partopsep=0pt,parsep=0pt,itemsep=0pt]
    \item Data transfer between devices may experience delays due to network congestion. As a result, some devices cannot get the information they require to finish a task as quickly as they should.
    \item Stragglers can also be caused by device faults. A gadget might crash or run out of battery life, preventing it from finishing a task.
    \item Devices deployed in challenged settings may be more likely to have restricted resources, such as memory or computing power. This is because the workload of an FL task may be too much for these devices to handle quickly.
\end{itemize}

One of the major roadblocks in FL is straggler parties, among the $S^{(r)}$. These parties stall the overall FL training as they do not communicate their models within the given time threshold for local training. This results in under-representing the straggler's data while aggregating the global model. 



\subsection{Security and Privacy of FL}~\label{sec:flsecpriv}

Recent research \cite{membinfsurvey,DLG-git, zhao2020idlg,geiping2020inverting, yin2021see, zhu2019deep} on reverse engineering attacks has demonstrated that while parties in federated learning (FL) do not share training data, sharing model updates may not ensure privacy.
Two common types of reverse engineering attacks are (i) Data reconstruction attacks and
(ii) Membership/Property inference attacks. Data reconstruction~\cite{geiping2020inverting, zhao2020idlg, zhu2019deep} aims to reconstruct most, if not
all of the original training data of a participant from the global model updates (gradients). 
Membership inference~\cite{shokri-membership-inference, membinfsurvey} aims to determine whether a specific data point was used to train the global model. This can be done by analyzing the performance of the global model on carefully crafted inputs. Property inference aims to infer certain properties of the participants' data, such as their location or demographics. This can be done by analyzing the global model updates or by training a new model to predict these properties from the global model updates. Empirically, these attacks work well 
only on gradients corresponding to individual data items or on aggregated gradients
 computed from a small number (e.g., up to batch sizes 
of 16 for \cite{geiping2020inverting, zhao2020idlg, zhu2019deep}) of data items. Therefore, individual model updates have to be private, and FL is not considered private when the number of
parties is less than three.

Secure aggregation~\cite{bonawitz2017practical,huba2022papaya,bonawitz2019towards} is a vital technique in federated learning (FL) that safeguards data privacy and security from individual devices or clients, facilitating collaborative model training. The primary objective is to preserve the privacy of individual updates while still allowing them to contribute to enhancing the global model. Four key secure aggregation techniques, which can be combined as needed for defense-in-depth, include: (i) Homomorphic Encryption (HE), (ii) Differential Privacy (DP), (iii) Secure Multi-Party Computation (SMPC), and (iv) Trustworthy Execution Environments (TEEs) or Secure Enclaves.

HE (e.g., \cite{paillier1999public, bgv}) for FL involves parties sharing
a common public/private keypair. Parties encrypt the model updates before transmitting them to the aggregator
which performs the aggregation computation on the encrypted data. Aggregated model updates can then 
be decrypted by the parties. HE does not change model utility, but is computationally expensive -- two or three orders of magnitude even with the use of specialized hardware~\cite{batchcrypt_he, heco, sok}, and also results in significant increase in the size
of the model update (e.g., ~64$\times$ for Paillier HE~\cite{batchcrypt_he} which is sufficient
for many FL algorithms). While HE is practical for FL in \emph{cross-silo} datacenter/cloud settings where its latency and bandwidth requirements can be accommodated, the need
for all participants to share a common keypair makes it impractical for large scale settings.

Differential privacy~\cite{abadi-diffpriv, kamalika-diff-private-sgd, nathalie-hybrid-eff, nathalie-hybridalpha} is a statistical technique that adds controlled noise 
to the model updates before aggregation. This noise ensures that individual updates do not reveal sensitive information about the data used for training, defeating data reconstruction and 
membership inference attacks. Some techniques also clip the model updates (gradients) to a predefined range before the addition of noise. This further limits the information that can be extracted from the updates. By carefully controlling the amount of noise, differential privacy provides a trade-off between privacy and utility. However, this is non-trivial, and model utility is very sensitive 
to an optimal choice of hyperparameters which is difficult in large-scale FL settings.

Secure Multi-Party Computation (SMPC) protocols~\cite{spdz-bmr-eurocrypt, damgard-spdz, spdz-csiro, jayaram-cloud2020, bonawitz2017practical} allow multiple parties (devices in this context) to perform computations on their inputs while keeping those inputs private. In FL, SMPC is used to aggregate model updates securely. Each party encrypts its update, and multiple parties perform computations on the encrypted updates without revealing the raw data. The result is an aggregated update that can be used to update the global model. Many SMPC protocols also suffer similar drawbacks as HE, 
with increased communication and computation time, lower in magnitude than HE but still significant,
and the need for effective key distribution.

A trusted execution environment (TEE)~\cite{amdsev,gravitron} is a 
secure area of a main processor that provides security features for isolated execution and guarantees the integrity of applications executing within, along with the confidentiality of their data assets. ARM TrustZone~\cite{trustzone} and Intel SGX~\cite{intelsgx} are examples of TEEs.
For aggregation in FL, they are attractive because their computational
overhead is low, and their computations and software can be audited by participants
with the help of attestation services. At least one FL system deployed at scale -- Papaya~\cite{huba2022papaya} uses TEEs for aggregation.

\section{\ours : Design}
\label{sec:flips}

Participant selection is a key challenge in FL. It is the process of choosing which devices will participate in each round of training, 
and is predominantly random~\cite{fedprox, fedma, bonawitz2019towards, fieldguide}. 
There is existing research on participant selection to optimize communication costs and computation 
limitations, which does not consider parties' data distribution and data diversity. 
For example, \cite{Huang_2020} models the client selection process as a Lyapunov optimization problem. 
The authors propose a C2MAB-based method to estimate the model exchange time between each client and the server. 
They then design an algorithm called RBCS-F to solve the problem. 
VF-PS~\cite{jiang2022vfps} is a framework for selecting important participants in vertical FL (VFL) efficiently and securely. It works by estimating the mutual information between each participant's data and the target variable and then selects the most important participants based on their scores. To ensure efficiency, VF-PS uses a group testing-based search procedure. To ensure security, it uses a secure aggregation protocol. VF-PS  achieves the target accuracy faster than training a naive VFL model.
FedMCCS~\cite{Abdulrahman2021FedMCCSMC} addresses challenges in using FL with IoT devices. FedMCCS considers the CPU, memory, energy, and time of the client devices to predict whether they are able to perform the FL task. In each round, FedMCCS maximizes the number of clients while considering their resources and capability to train and send the needed updates successfully. FedMCCS outperforms other approaches by reducing the number of communication rounds to reach the intended accuracy, maximizing the number of clients, ensuring the least number of discarded rounds
and optimizing the network traffic. \cite{Nishio_2019} takes a similar approach as FedMCCS, 
prioritizing resource availability.
Another approach considers data valuation for compensating valuable data owners~\cite{principled_data_valuation}. They use the Shapley value as a fair allocation mechanism that assigns a value to each data source based on its contribution to the model's performance to enhance system robustness, security, and efficiency; such mechanisms could be used for aiding participant selection.
\cite{cho2021client} proposes Power-of-Choice, a communication- and computation-efficient client selection framework, which randomly selects 
a fraction of participants in each round, but biases selection towards those with higher
local losses. \cite{cho2021client} proves theoretically and empirically that this biased 
selection leads to faster convergence. Oort~\cite{Oort-osdi21} takes a similar
approach. 

In this section, we introduce \ours, our approach to intelligent participant selection that improves model 
convergence, addresses diversity in datasets and incurs low communication overheads. 
First, \ours\ mitigates the above-mentioned class imbalance issue,
(Algorithm~\ref{alg:fed:flips}) and  improves feature and participant representation in FL 
by selecting parties that are likely to have
dissimilar data at each FL round. The techniques implemented within \ours\ are 
based on one of the federated
learning's core goals -- to increase the diversity of data and 
ensure that the global model in each FL round does not overfit local data.

\subsection{Finding Similar Parties using Clustering}

The objective of FLIPS is to identify sets of similar parties by measuring the label distribution of each party and using it as a semantic representation of a party's local dataset. There are $N$ parties, $p_1, p_2, \ldots, p_N$, with datasets $d_1, d_2, \ldots, d_N$ and label distributions $ld_{1}, ld_{2}, \ldots, ld_{N}$, respectively.
The label distribution vector at party $p_i$ with dataset $d_i$ is 
$ld_i = \{l_1, l_2, \ldots, l_g\}$, where $l_j$ is the number of data points for the label $j^{th}$ present in the party and $g$ is the number of labels in the dataset. The set of label distributions for all $N$ parties is denoted as $LD = \{ld_{1}, ld_{2}, \ldots, ld_{N}\}$.

Our next step is to group the label distributions from various parties into non-intersecting subsets that are similar. Here, we define a similarity metric between subsets that is based on the average distance between all objects in a given subset and the average distance between subsets. Let $S_m$ be the set of all possible subsets of $LD$ of size $m$, where $m~\epsilon~[1,N]$.
Hence, there are $N \choose m$ subsets within $S_{m}$. Let $L_{i} = \{ld_{a},ld_{b},....\},$ be a subset $\epsilon~S_{m}$, $\Delta(L_{i})$ is the average Euclidean distance between all objects in the set $L_{i}$. Given 2 subsets $L_{i}$ and $L_{j}$, $\delta(L_{i},L_{j})$ is the average Euclidean distance between objects in sets $L_{i}$ and those in $L_{j}$.

The idea behind finding similar parties is to find $k$ disjoint subsets across all $S_m$, where $m~\epsilon~[1,N]$, such that: 

\begin{equation}
  minimize_{S_{m}}\frac{1}{k}~\sum_{i=1}^k~\sum_{j=1}^k \frac{\Delta(L_{i}) + \Delta(L_{j})}{\delta(L_{i},L_{j})}, i\neq~j. 
\end{equation}

Note that this problem is a subset enumeration problem, where we have to find $k$ subsets across all $S_{m}$, where the condition is (1). This problem is known to be NP-complete~\cite{doi:10.1137/0208032}. There are several heuristics to solve this problem, we use K-Means~\cite{1056489} clustering to solve this problem which obtains a $k$-partition, where $k$ is unknown, across all $S_{m}$, denoted by clusters $C = (C_{1}, C_{2}, \ldots, C_{k})$, such that :
\begin{equation}
\begin{aligned}
& \text{minimize} && \sum_{x=1}^N \sum_{y=1}^k \omega_{xy} ||ld_x - c_y||^2 \\
\end{aligned}
\end{equation}

Here, $\omega_{xy}$ is a binary variable that indicates whether the $x$th datapoint is assigned to the cluster $C_{y}$, whose centroid is $c_{y}$.

\begin{figure}[h]
    \centering
    \includegraphics[width=0.3\columnwidth]{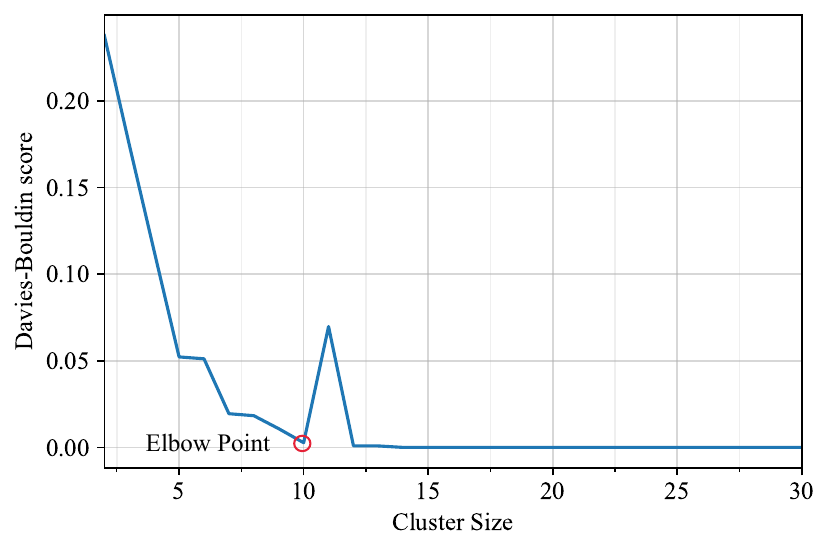}
    \caption{Elbow point determination for  optimal \textit{k}}
    \label{fig:elbow:point}
\end{figure}

We opted for K-Means due to its simplicity and lower time complexity. K-Means clustering has a time complexity of $O(NkI*d)$, where $N$ is the number of data points, $k$ is the number of clusters, $I$ is the number of iterations, and $d$ is the number of dimensions. This makes it a suitable choice for resource-limited settings. Furthermore, we use \textit{ kmeans ++}~\cite{10.5555/1283383.1283494} to initialize the centroids in K-Means clustering. This has been demonstrated to scale
to millions of data points, i.e., parties~\cite{kmeansscalability}.
However, K-means does require the number of clusters $k$ to be known beforehand, 
which is a problem in FL.
The knowledge about the number of unique label distributions in the parties' datasets is unknown while performing clustering, as each party's data is kept private. To find the optimal number of clusters $k$, which is analogous to the number of unique label distributions, we use a purity metric called the Davies-Bouldin index ($dbi$)~\cite{4766909}, which is the ratio of the intra-cluster distance to the inter-cluster distance and similar to (1). The optimal k is determined by :
\begin{equation}
    k_{optimal} = arg min_{k} |\frac{dbi(k) - dbi(k-1)}{dbi(k-1)} |.
\end{equation}
 where dbi is the Davies-Bouldin index.

When $k$ is small, the clusters cannot accurately represent the unique label distributions, impacting FL performance and cost because there is no equitable representation at each round. When $k$ is large, clustering leads to overfitting; clusters generated are sparse, impacting FL performance and cost. To determine the optimal cluster size in \ours\ \emph{empirically}, we experiment with different cluster sizes in succession, repeated $T=20$ number of times (because K-Means is sensitive to the centroid initialization) and average the $dbi$ for each $k \in \{2, \ldots, K\}$, where $K=N$. This gives us $T$ different $dbi$ for each cluster size $k$. The cluster size $k$ for which there is a (first) sharp change in the slope of the curve (elbow point) is chosen as the optimal cluster size. As illustrated in Figure~\ref{fig:elbow:point}, the optimal cluster size at which there is a sharp change in slope for $k$ vs $dbi$ is 10. Hence, we choose 10 as the cluster size for K-Means.

\subsection{Intelligent Participant selection}

Given a set of clusters of parties, $\mathcal{C}$, from the clustering technique described above, \ours\ (Algorithm~\ref{alg:fed:flips}) implements participant selection for a round by 
choosing one party at a time from each cluster in a round-robin manner until the  number of parties required for the round, $N_r$, is reached.  Typically in FL training, 
the number of parties per round $N_r$ is fixed across all the rounds. This ensures that 
$N_r$ is spread among as many clusters as possible increasing the diversity of data in 
the FL training process. It is  recommended that $N_r$ be a multiple of 
the number of clusters $|\mathcal{C}|$ since $N_r$ can then be easily split among the number of available clusters ($|\mathcal{C}|$), ensuring equal 
representation from each cluster ($\frac{N_r}{|\mathcal{C}|}$). 
\ours\ also keeps track of the number of times a party was chosen to ensure that  each party within a cluster is given an equal opportunity to participate. 
In the case where the number of parties per round is less than the number of clusters, 
not every cluster can participate in every round. So \ours\ additionally tracks
the number of times a cluster is selected to participate. 

Consider the example of using ECG signals from wearables for Arrhythmia detection, 
\ours~will improve label representation by picking parties with normal and abnormal 
heartbeats at each round, improving the detection rate for 
arrhythmia and preventing class/label imbalance. To improve participant 
representation, \ours~will try to incorporate parties that were not used in the 
previous training rounds. This will help bring knowledge from a diverse 
set of participants and will make the global model more robust. This helps solve the data heterogeneity issues in FL.



To mitigate the effect of Stragglers, \ours~uses the popularly used over-provisioning technique~\cite{tff}. Once we identify the average straggler rate $strg$. \ours~overprovisions $strg*S^{(r)} $ parties in the subsequent training rounds. The overprovisioned parties in round $r+1$ are selected from the clusters $H_{sc}^{r}$ that the straggler parties in round $r$ were a part of. In this manner, we maintain the representation of all the unique label distributions in FL. This is illustrated in Algorithm~\ref{alg:fed:flips}. 




\begin{algorithm}
\caption{\ours~Party \& Straggler Handling}~\label{alg:fed:flips}
\mbox{{\bf Participant Side}} \\ 
    
    \textsc{recv} \changes{global model} ($m^{(r)}$) from aggregator \\
    Local model $x^{(r,1)} \gets m^{(r)}$\\
    \For(){$k \in \{ 1,2,\ldots,\tau \} $}  {  \tcp{$\tau$ local iterations}
          Compute local stochastic gradient $g_i(x^{(r,k)})$\\
          $x^{(r,k+1)} \gets \textsc{optimizer}(x^{{r,k}}, -g_i(x^{(r,k)}), \eta^{(r)})$\\
    }
    
    \textsc{send} $x^{(r,\tau)}$ to aggregator \\

\mbox{{\bf Aggregator using \ours}} \\ 
        
    $LD = \{ld_{1}, ld_{2}, \ldots, ld_{n}\} $ \\
        
    $C \gets \textsc{optimized\_clusters(LD)}$\\

    $H \gets \{\}$ \tcp{parties used per cluster} 
    $H_{c} \gets  \textsc{min-heap()}$ \tcp{clusters used}
    $H_{s}^{r}\gets \{\},~~H_{sc}^{r} \gets $\textsc{max-heap()},~$count\_strg~\gets~0,  Stragglers=False $ \tcp{track straggler parties, their clusters and counts} 
        
    $\mbox{Initial model } m^{1}, \mbox{Parties per round}~N_{r}$ \\

    \For(){$c \in \{1,2,\ldots,C \} $ } { 
        $c.picks \gets 0,~~\textsc{insert}(H_c, c),~~h \gets \textsc{min-heap()}$\\
        \For(){$p \in \{1,2 \ldots c\}$}{
            $p.picks \gets 0, \textsc{insert}(h,p)$\\
        }
        $H[c.id] \gets h$\\
    }
    \For(){$r ~\in~ \{1,2,\ldots,R\}$}{
             $S^{(r)} \gets \{\}$ \\
        \For(){$i ~\in~ \{1,2,\ldots,N_{r}\}$}{

            $c \gets \textsc{extract\_min}(H_{c}),~~H \gets H[c.id],~~p \gets \textsc{extract\_min}(H)$ \\
            $\textsc{increement}(p.picks,1),~~\textsc{insert}(H,p)$ \\
            $\textsc{increement}(c.picks,1),~~\textsc{insert}(H_{c},c)$ \\
            Select unique parties :$S^{(r)} \gets S^{(r)}\bigcup \{p\} $ \\
        }

        \If{Stragglers}{
            \For(){ $i ~\in~ \{1,2,\ldots,int(strg*N^{r})\}$}{
                $c\gets\textsc{extract\_max}(H_{sc}^{r})$, $H \gets H[c.id]$  \tcp{choose cluster with most stragglers}    
                $c$ : $p \gets \textsc{extract\_min}(H)~not~in~H_{s}^{r}$ \tcp{pick a non-straggler part in $c$}
                $S^{(r)} \gets S^{(r)}\bigcup \{p\} $\\
            }
        }
      
        \textsc{send} $m^{(r)}$ \mbox{to each} $i \in  \mathcal{S}^{(r)}$ \\
        
        \For(){\textsc{recv} \mbox{model update} $x_{i}^{(r)}$ \mbox{from each} $i \in  \mathcal{S}^{(r)}$} {
            
            \If{$x_{i}^{r}$ not \textsc{recv}} {
            
                $H_{s}^{r}\gets i$, $H_{sc}^{r}\gets$ cluster of i \\
            
                \textit{Stragglers=True}, $count\_strg++$  \\
            }
            \Else{
                
                \If{$i~in~H_{s}^{r}$}{
                    $H_{s}^{r}.remove(i)$\\
                    $H_{sc}^{r}.remove(cluster~of~i)$
                }
            
            }
        }

        \If{$len(H_{s}^{r})==0$} 
        {
            $Stragglers=False$
        }
            
        {\mbox{Aggregate } $x^{(r)} \gets \frac{1}{N} {\sum_{i \in  \mathcal{S}^{(r)}} n_i x_{i}^{(r)}}$ }\\
            
        $m^{(r+1)} \gets \textsc{optimizer}(m^{r},x_{i}^{(r)})$ \\

        $strg = \frac{strg *N^{r} + count\_strg}{N^{r}}$
    }      

\end{algorithm}

\subsection{Privacy in \ours}

There are two additional pieces of private information that needs to be secured in \ours,
when compared to traditional FL (see Section~\ref{sec:flsecpriv}). One is the parties' label distribution
used for clustering. This information gives away the types of data present at a party, and
can be used by other parties to gain a competitive advantage. The other is cluster membership -- should
a party know which cluster it belongs to, Attacks on
FL algorithms based on cluster membership are currently unknown, given that the concept and use in FL is
new. But, akin to data and model poisoning, parties could lie about their label distributions
to work themselves into smaller clusters. In \ours, we treat cluster membership as private information
because parties \emph{do not need to know this.} A party simply needs to know whether it is selected
for a round.

Hence, \ours\ needs a mechanism to establish trust and execute clustering privately and securely.
For this, we have chosen to use secure enclaves (Section~\ref{sec:flsecpriv}) over 
secure multiparty computation (SMPC) and homomorphic encryption (HE).Figure~\ref{fig:sys:design:tee} illustrates the
end-to-end system design of \ours\ and the associated flow of information.  The key components include the FL parties which hold local data and generate local models, an aggregator which acts as the centralized coordinator, the TEE within the centralized aggregator, and an attestation server that verifies the aggregator's TEE.
The clustering code is loaded on the TEE and all parties share an attestation server to verify the remote TEE's hardware. Together, this establishes the TEE as a secure enclave. 

Next, each party establishes a secure channel (eg: TLS channel) with the TEE for transmitting secrets,  the label distribution vector in our case. The clustering code (a K-means variant) computes the clusters on these label distribution vectors in the secure enclave.  The TEE maintains the clusters securely and deletes all information at the end of the FL job (this can
be attested by the attestation server).

\begin{figure}[t!]
\centering
\includegraphics[width=0.50\columnwidth]{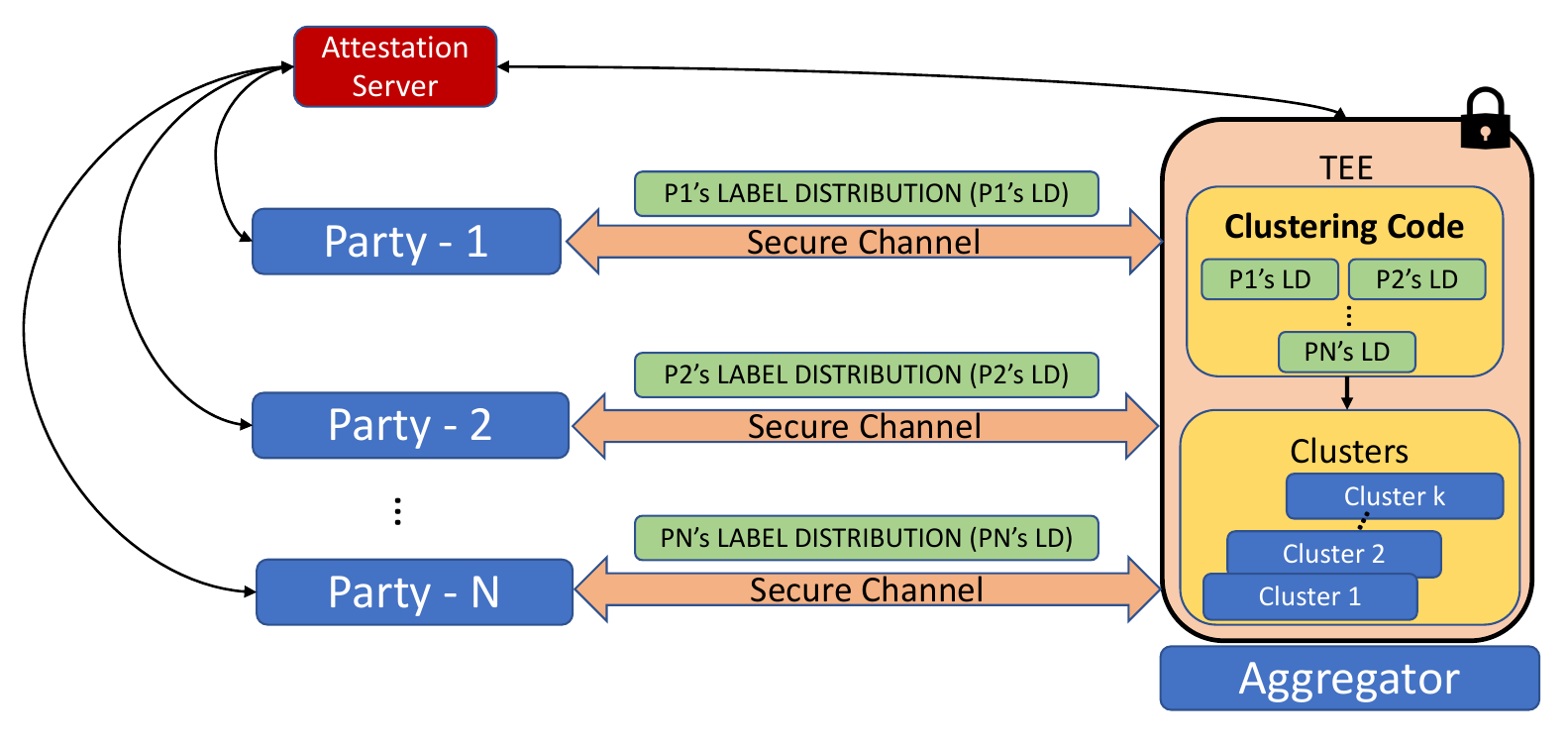}

  \caption{End-to-end integrated system design for Private Clustering in \ours
  }
    \label{fig:sys:design:tee}
\end{figure}

\begin{figure}[t!]
\centering
\includegraphics[width=0.50\columnwidth]{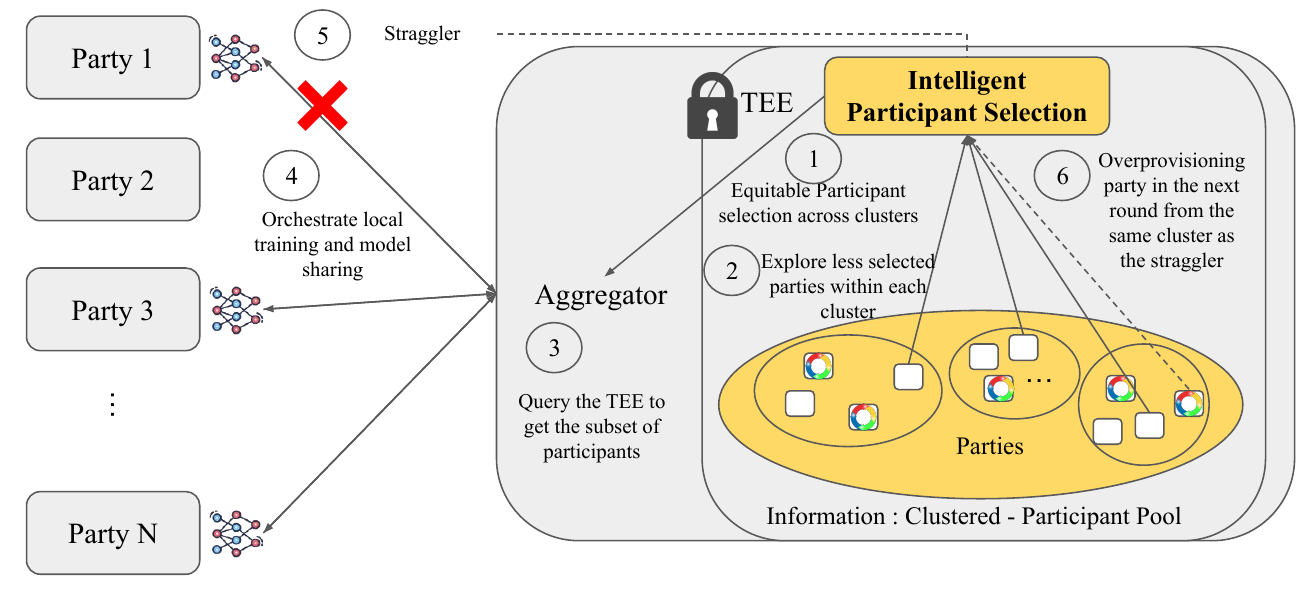}
  \caption{Workflow: Intelligent Participant Selection in  \ours
  }
    \label{fig:sys:ips:workflow}
\end{figure}

\subsection{\ours\ in Context}

\ours\ differs from other FL systems, in that it targets participant selection using label distribution
clustering. k-means++ clustering based on label distributions has the advantage of being fast and
minuscule relative to FL training time, and has been demonstrated to scale~\cite{kmeansscalability, marigold} 
to millions of data points (participants). Clustering has to be performed only once, as long as the
set of participants or the data at participants does not change significantly. 
Centralizing this clustering and participant selection in a TEE also has the advantage of fitting
nicely into the predominant mold of FL, as it is deployed today, based on a single aggregator~\cite{kairouz2019advances}.
However, \ours\ can also be used with a distributed aggregator (e.g., ~\cite{bonawitz2019towards,huba2022papaya,jkrbigdata,jkrmascots}) by separating the clustering and participant 
selection module from the aggregation module. This applies to cloud-hosted aggregation or
aggregation using homomorphic encryption or secure multiparty computation. In fact, clustering, participant selection
and aggregation are logically separate with clear interfaces, and can be individually hosted and secured.
This also implies that \ours\ is as scalable as the underlying aggregation algorithm and the 
method used to secure it.

\section{Validating \ours}

Our primary goal is to examine the impact of \ours's participant selection
on model convergence and accuracy at a reasonable scale. Hence, we implemented and 
evaluated \ours~in a distributed cluster environment. 
All the parties, i.e. participants, were executed as nodes on a cluster for local 
training as seen in Figure~\ref{fig:sys:design:tee}. 
The cluster consists 13 nodes, each with four Nvidia V100 (16GB) GPUs, Intel(R) Xeon(R) 
Silver 4114 CPU, and $\sim$ 16 GB RAM per node.  We train local models using the GPUs 
available on this cluster. 
The aggregator node is executed on a machine with a 2.9Ghz 6-core Ryzen 5 processor with 16GB RAM and 512GB SSD.
To enable trusted execution for label distribution clustering, we use the 
AMD Secure Encrypted Virtualization (SEV)~\cite{amdsev} on the aggregator. 

\subsection{Techniques for Comparison}

We compare \ours\ with 3 popular participant selection strategies across different datasets 
and study the impact on convergence in the presence of stragglers. 
The first is the widely-used  
random selection~\cite{mcmahan2017communication,fedprox,https://doi.org/10.48550/arxiv.2003.00295} method. 
This selects each party with the same probability and can lead to class imbalance, where certain classes are underrepresented, as explained in Section~\ref{sec:randomselectionproblems}. 
This may cause the model to be biased towards the overrepresented classes, leading to poor performance. 
This harms convergence as it takes more training rounds to reach the desired accuracy.

Second, we compare \ours\ with OORT~\cite{Oort-osdi21}. OORT uses the idea that parties with a higher 
local loss can contribute more to an FL job~\cite{pmlr-v151-jee-cho22a} and introduces a 
statistical and systemic \emph{utility metric} for participant selection. The system sorts the 
parties according to the party utility metric which is a combination of its statistical and systemic 
utility and selects parties with a higher utility and explores new parties at each training round. 

The third strategy is GradClus~\cite{pmlr-v139-fraboni21a}, which uses the idea of clustering gradients from parties to 
identify parties with similar data. It performs hierarchical clustering over a similarity 
matrix constructed across gradients from all the parties in the FL job. The gradients assigned in 
the beginning are random numbers and get iteratively updated as the party gets picked. 
At the aggregator, GradClus~\cite{pmlr-v139-fraboni21a} performs hierarchical clustering and $S^{(r)}$ number 
of clusters are formed. GradClus chooses one party from each cluster randomly. 

The fourth strategy is TiFL~\cite{tifl}, which groups parties into tiers based on their training performance and selects parties from the same tier in each training round to mitigate the straggler problem. To further solve the non-IID problem, TiFL employs an adaptive tier selection approach to update the tiering on the fly based on the observed training performance and accuracy.

\subsection{Datasets and models used}

We focus on two real-world datasets from the healthcare/senior care domain and two benchmark datasets:

\begin{itemize}[topsep=0pt,partopsep=0pt,parsep=0pt,itemsep=0pt]
    \item MIT-BIH-ECG-Dataset~\cite{932724} partitioned across 200 parties trained using a 1-D CNN~\cite{9797945}.
    \item Skin cancer HAM10000~\cite{codella2019skin} partitioned across 200 parties trained using Densenet-121~\cite{huang2018densely}.
    \item FEMNIST~\cite{fedprox} partitioned across 200 parties trained using Le-Net5~\cite{lenet}.
    \item FashionMNIST~\cite{xiao2017/online} partitioned across 200 parties trained using Le-Net-5~\cite{lenet}.
\end{itemize}


\textbf{The MIT-BIH ECG dataset}~\cite{932724} comprises digitized electrocardiogram (ECG) recordings used for arrhythmia identification. Collected by MIT Biomedical Engineering Department and Beth Israel Hospital, it includes both normal and aberrant rhythms. The dataset is annotated with AAMI labels~\cite{6713540}, a widely accepted standard for ECG rhythm classification. These labels define performance criteria, improve algorithm generalization, and include N (normal beats), S(supra-ventricular ectopic beats), V (ventricular ectopic beats), F (fusion beats), and Q (unclassifiable beats). The dataset predominantly comprises of N beats, necessitating federated learning (FL) to enhance label and participant representation. The dataset is distributed across 200 parties in a non-IID manner. Training involves a 1-D CNN with a learning rate 0.001 and a decay applied every 20 rounds. FL training is limited to a maximum of 400 rounds.

The \textbf{HAM10000} dataset~\cite{codella2019skin} contains diverse dermatoscopic images of pigmented skin lesions. It includes 10015 images representing important skin cancer diagnostic categories: akiec, bcc, bkl, df, mel, nv, and vasc. The dataset is suitable for training and evaluating machine learning models for automated diagnosis. The nv images are the most abundant, potentially dominating other categories due to their prevalence caused by UV radiation. This non-IID behavior highlights the need for federated learning. The dataset is distributed across 200 parties, and training involves DenseNet121 with a learning rate of 0.001. A decay is applied every 30 FL rounds, and the maximum number of FL rounds is 400.

The \textbf{EMNIST} (Extended MNIST) dataset~\cite{https://doi.org/10.48550/arxiv.1702.05373} contains handwritten letters and numbers and is an expansion of the MNIST dataset. The EMNIST dataset contains characters from numerous alphabets, including numerals and letters from the English alphabet. It also has a collection of symbols from different languages. This dataset is frequently used for testing and training machine learning models for tasks like text classification and handwriting recognition.
We subsample 10 lowercase characters ('a'-'j') from EMNIST. This federated variant of EMNIST is known as FEMNIST~\cite{fedprox}. We train Le-Net-5~\cite{lenet} model at each party which outputs a class label between 0 (a) and 9 (j).

\textbf{Fashion-MNIST}~\cite{xiao2017/online} is a dataset of images of clothing items that are commonly used as a benchmark for machine learning models. It consists of 60,000 training and 10,000 test images, each 28x28 pixels in size and labeled with one of 10 different clothing categories, such as t-shirts, trousers, bags, etc. FL applied in the fashion-MNIST dataset will mimic a personalized customer recommendation system. A model trained on the customer's device or organization using FL will understand the customer's preferences, which can then be used to suggest personalized clothing recommendations. The dataset is distributed across 100 parties and training involves Le-Net5~\cite{lenet}.

\subsection{Emulating Non-IID data distributions}~\label{sec:alphaexplanation}

As in the Tensorflow Federated~\cite{tff} and LEAF~\cite{caldas2018leaf} FL benchmarks, we 
emulate a non-IID setting in our experiments by using different data partitioning strategies. We use Dirichlet Allocation~\cite{https://doi.org/10.48550/arxiv.1905.12022} a widely
used technique~\cite{tff, caldas2018leaf} to partition a dataset among several
parties in a non-IID manner. It samples $p \sim Dir_{N} (\alpha)$, where $\alpha$ is the 
control parameter and $p_{l,i}$ becomes the proportion of the number of 
data points of label $l$ allocated to the party $i$. An $\alpha$ of 0 corresponds
to each party having data corresponding to only one label (which is non-IID at its extreme)
and an $\alpha ~\geq 1$ corresponds to an IID distribution.
As recommended in other federated learning research~\cite{gao2022federated, tff-benchmark, caldas2018leaf}, we evaluate \ours\ using two different values of $\alpha=0.3$ and $0.6$.
At each round, we sample 20\% and 15\% parties which is more than the number of clusters of similar parties formed in \ours.

\subsection{Metrics}

In FL, datasets are local to the parties, which partition them into local training and test sets.
To compare \ours~and other techniques for this study, we use a global test set consisting of data 
corresponding to all the labels in the distributed dataset. 
The datapoints in this dataset are unknown to any party in our emulation. Hence it is a valid test set. This test set also helps us evaluate the techniques at each training round to get a closer look at convergence. 
Generally, global test sets are not used in FL practice; just while designing FL algorithms.
We maintain the global test set inside the aggregator's TEE in our implementation. 

We report the test accuracy of the global model against the test dataset after each round. This test accuracy is computed as:

    \begin{align*}
      Acc & =  \frac{lA_{1} + lA_{2} \ldots + lA_{m}}{m} \\
      lA_{i} & =  \frac{Correct~Predictions~for~label~i}{Total~number~of~datapoints~for~label~i}
\end{align*}

This is done to mitigate the effect of label imbalance while computing the accuracy for each test set, as each label may have a different number of datapoints.
The accuracy numbers reported are an average of 6 runs for each experiment. 

We report both (i) the highest accuracy obtained using a specific FL technique within the FL rounds threshold and
(ii) the number of communication rounds needed for the global model to reach a target/desired accuracy.
The latter depicts how fast any participant selection technique converges.

\section{Results}~\label{sec:eval}

\begin{figure*}[h]
\centering
\subfloat[ $\alpha=0.3$ without stragglers]{{\includegraphics[width=0.45\textwidth]{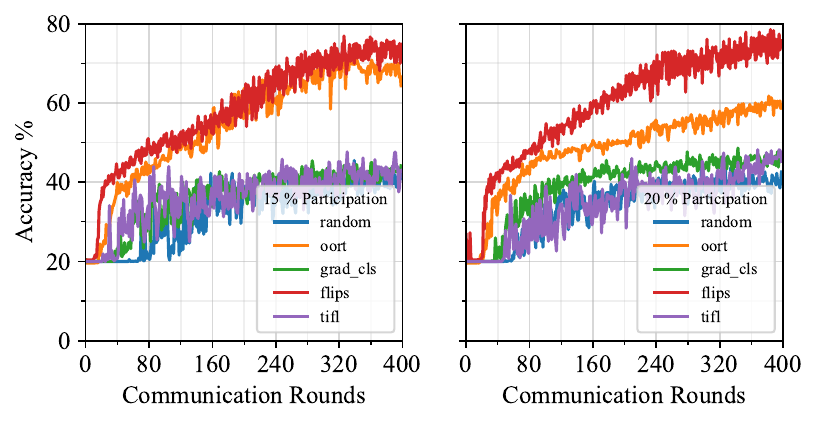}}%
    \label{fig:ecg:03}
    }
    \quad
    \subfloat[$\alpha=0.6$ without stragglers]{\includegraphics[width=0.45\textwidth] {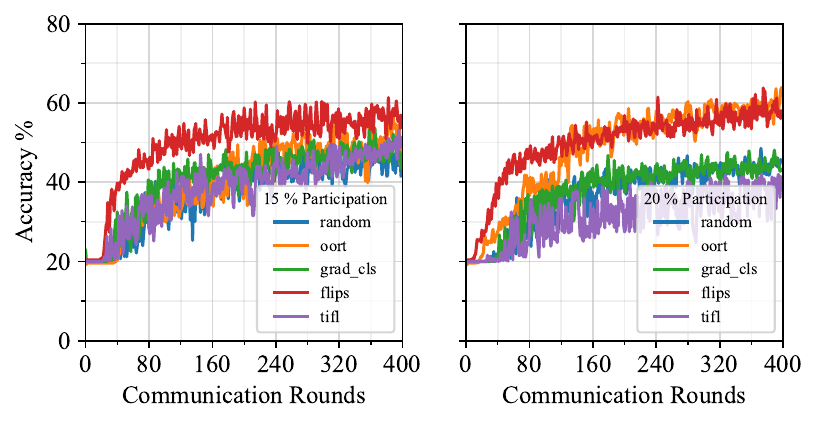}%
    \label{fig:ecg:06}
    }

  \caption{Convergence plots on MIT BIH ECG Dataset, FL Algorithm: FedYoGi}
  \label{fig:ecg:all}\label{fig:mit:strg:all}
\end{figure*}

\begin{figure*}[h]
\centering
\subfloat[$\alpha=0.3$ with stragglers]{{\includegraphics[width=0.45\textwidth]{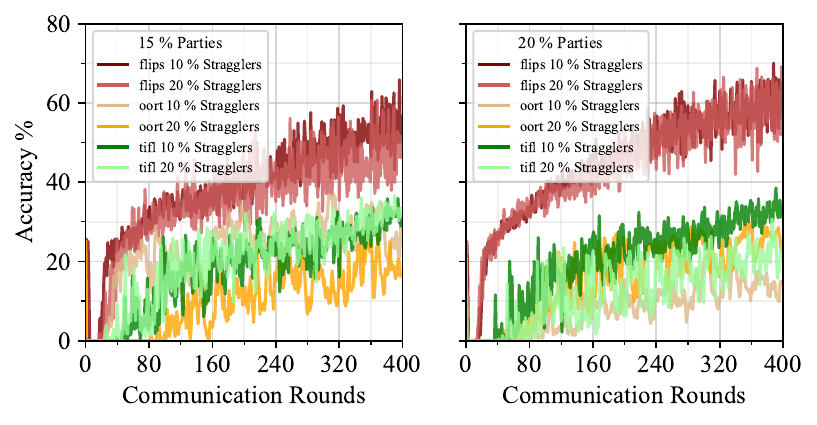}}%
    \label{fig:mit:strg:03}
    }%
    \quad
    \subfloat[$\alpha=0.6$ with stragglers]{\includegraphics[width=0.45\textwidth] {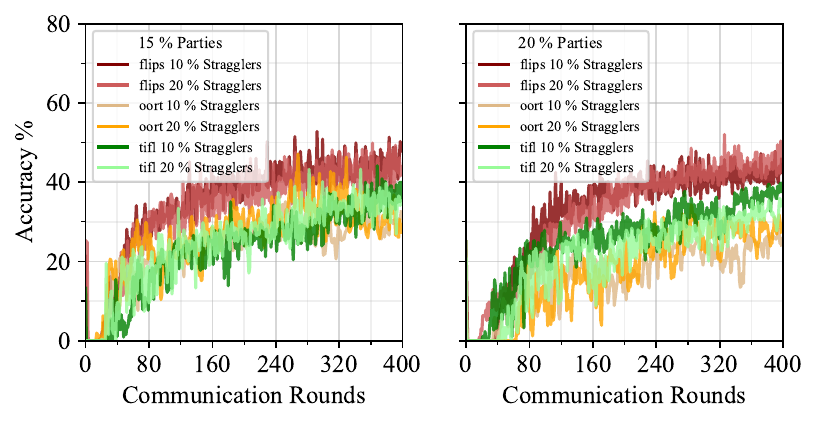}%
    \label{fig:mit:strg:06}
    }%
  
  \caption{Convergence plots on MIT BIH ECG Dataset in the presence of Stragglers, FL Algorithm: FedYoGi}
  \label{fig:mit:strg:all}
\end{figure*}

\begin{figure*}[h]
\centering
\subfloat[$\alpha=0.3$ without stragglers]{{\includegraphics[width=0.45\textwidth]{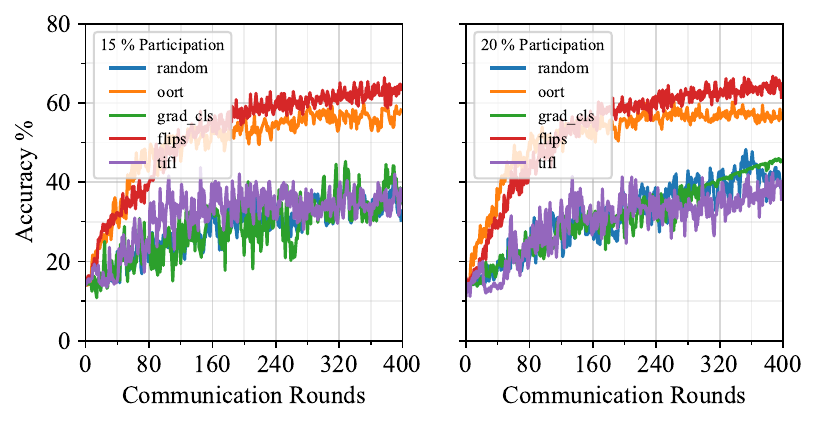}}%
    \label{fig:sk:03}
    }
    \quad
    \subfloat[$\alpha=0.6$ without stragglers]{\includegraphics[width=0.45\textwidth] {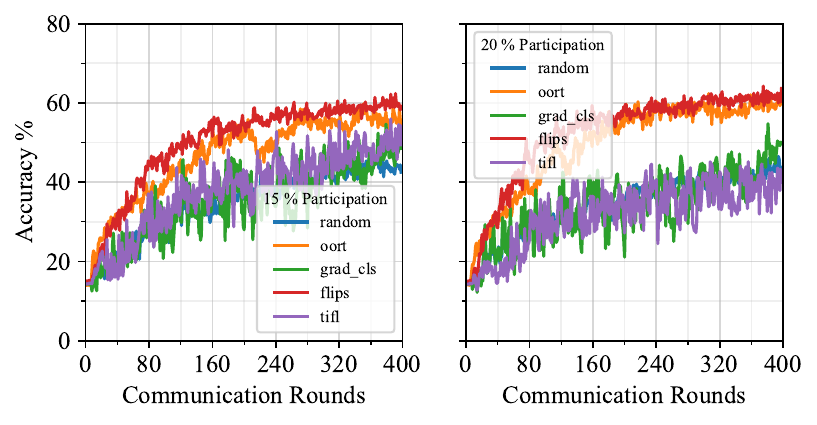}%
    \label{fig:sk:06}
    }

  \caption{Convergence plots on HAM 10000 (Skin Lesions) dataset, FL Algorithm: FedYoGi}
  \label{fig:sk:all}\label{fig:sk:strg:all}
\end{figure*}

\begin{figure*}[h]
\centering
\subfloat[$\alpha=0.3$ with stragglers]{{\includegraphics[width=0.45\textwidth]{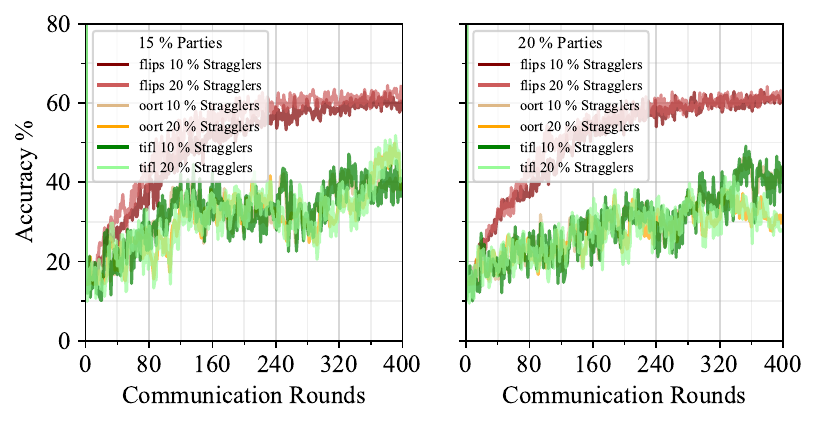}}%
    \label{fig:sk:strg:03}
    }%
  \quad
    \subfloat[$\alpha=0.6$ with stragglers]{\includegraphics[width=0.45\textwidth] {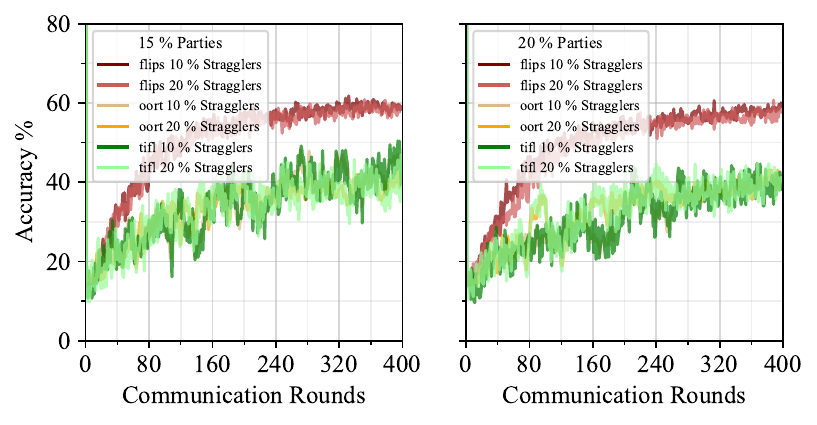}%
    \label{fig:sk:strg:06}
    }%
  
  \caption{Convergence plots on HAM10000 (Skin Lesions) Dataset in the presence of Stragglers, FL Algorithm: FedYoGi}
  \label{fig:sk:strg:all}
\end{figure*}

\begin{figure*}[h]
\centering
\subfloat[$\alpha=0.3$ without stragglers]{{\includegraphics[width=0.45\textwidth]{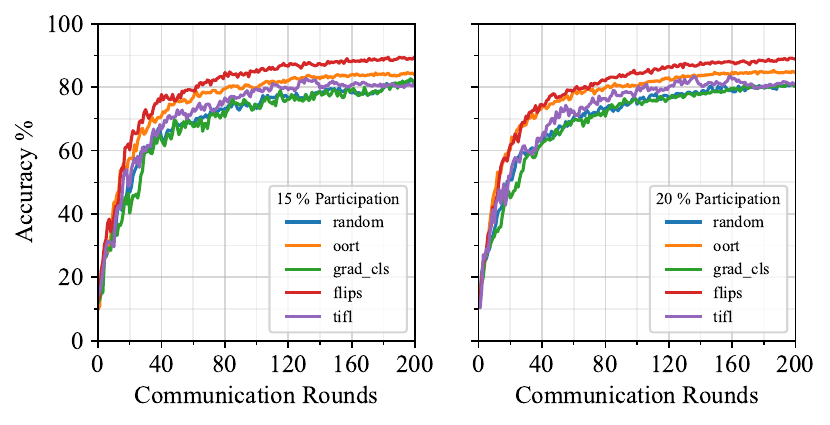}}%
    \label{fig:femnist:03}
    }
    \quad
    \subfloat[$\alpha=0.6$ without stragglers]{\includegraphics[width=0.45\textwidth] {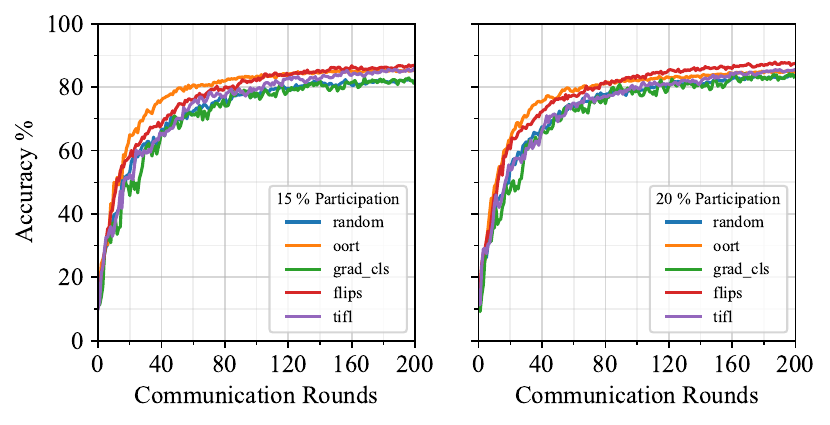}%
    \label{fig:femnist:06}
    }

  \caption{Convergence plots on FEMNIST dataset, FL Algorithm: FedYoGi}
  \label{fig:femnist:all}
\end{figure*}

\begin{figure*}[h]
\centering
\subfloat[$\alpha=0.3$]{{\includegraphics[width=0.45\textwidth]{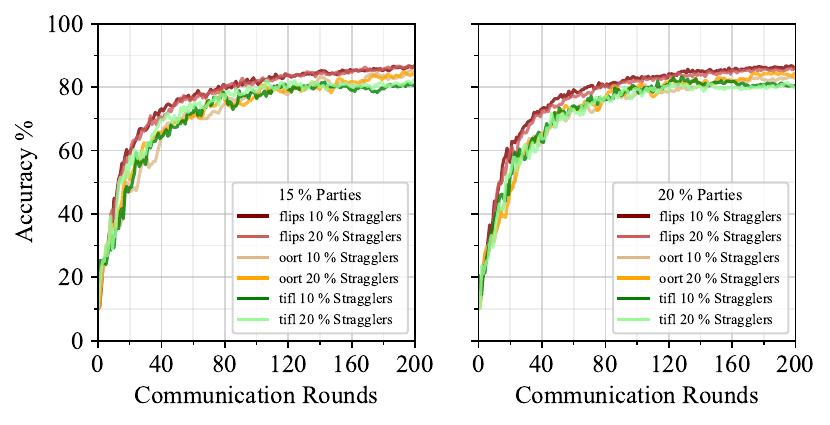}}%
    \label{fig:femnist:strg::03}
    }
    \quad
    \subfloat[$\alpha=0.6$]{\includegraphics[width=0.45\textwidth] {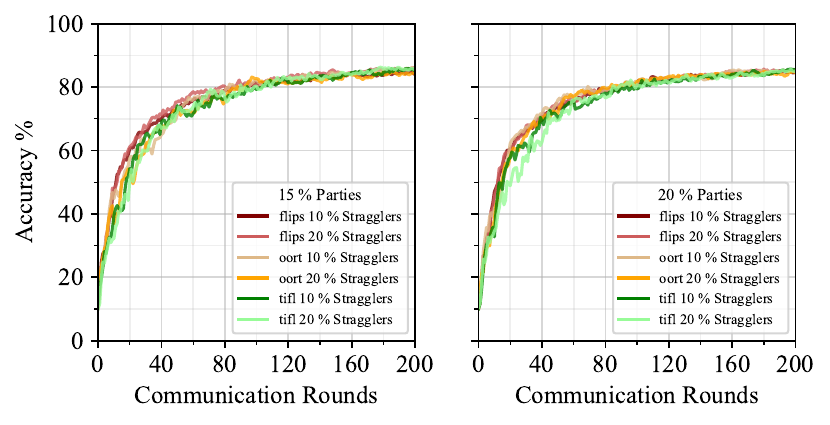}%
    \label{fig:femnist:strg:06}
    }

  \caption{Convergence plots on FEMNIST dataset in the presence of stragglers, FL Algorithm: FedYoGi}
  \label{fig:femnist:all}
\end{figure*}

\begin{figure*}[h]
\centering
\subfloat[$\alpha=0.3$ without stragglers]{{\includegraphics[width=0.45\textwidth]{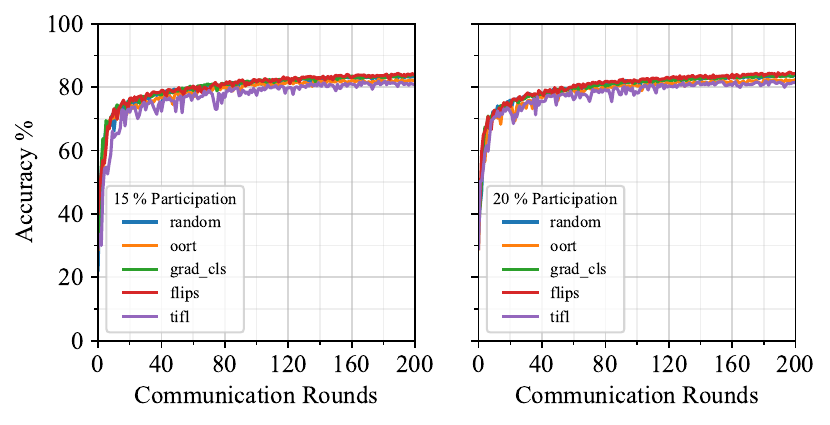}}%
    \label{fig:fashion:03}
    }
    \quad
    \subfloat[$\alpha=0.6$ without stragglers]{\includegraphics[width=0.45\textwidth] {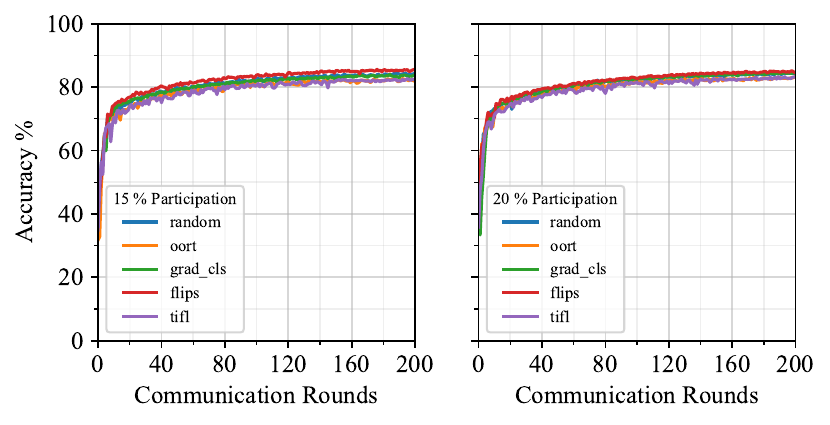}%
    \label{fig:fashion:06}
    }

  \caption{Convergence plots on Fashion MNIST dataset, FL Algorithm: FedYoGi}
  \label{fig:femnist:all}
\end{figure*}

\begin{figure*}[h]
\centering
\subfloat[$\alpha=0.3$]{{\includegraphics[width=0.45\textwidth]{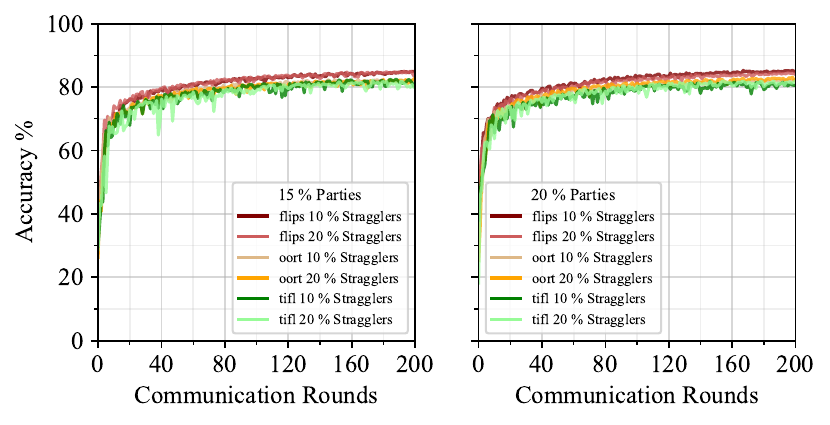}}%
    \label{fig:femnist:strg::03}
    }
    \quad
    \subfloat[$\alpha=0.6$]{\includegraphics[width=0.45\textwidth] {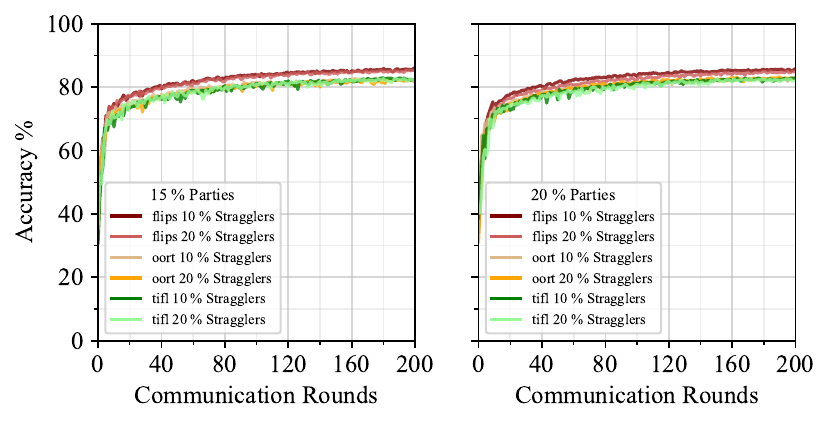}%
    \label{fig:fashion:strg:06}
    }

  \caption{Convergence plots on Fashion MNIST dataset in the presence of stragglers, FL Algorithm: FedYoGi}
  \label{fig:fashion:all}
\end{figure*}

We  
our results in this paper Tables~\ref{mit:rounds}--\ref{fashion:avg:accuracy}. 
Without stragglers, we perform a total of 60 ($5 \times 3 \times 2 \times 2 $) experiments --
comparing \ours\ with 4 other participant selection mechanisms (random, GradClus,
Oort, TiFL), for 3 different FL algorithms (\textit{FedAvg}, \textit{FedProx} and \textit{FedYoGi}), 
2 levels of non-IIDness ($\alpha=0.3$ and $\alpha=0.6$) and 2 levels of participation (20\% and 15\%) per dataset.
From these 60 experiments, we choose the 3 best-performing techniques \ours\ , OORT and TiFL and observe how they perform 
in the presence of stragglers.
We emulate stragglers by dropping 10\% or 20\% of participants involved in an FL round, resulting in another
72 experiments per dataset. 

\captionsetup[table]{skip=0pt}

\begin{table*}[]
\footnotesize
\centering
\caption{MIT ECG Dataset : Rounds required to attain Target Accuracy (60\%)}

\begin{tabular}{cc|ccccccccccc}
\hline
\multicolumn{2}{c|}{\multirow{2}{*}{\textbf{Setting}}} & \multicolumn{11}{c}{\textbf{Rounds required to attain 60 \% accuracy, FL Algorithm : FedYoGi}}                                                                                                                                                                      \\ \cline{3-13} 
\multicolumn{2}{c|}{}                                  & \multicolumn{5}{c|}{\textbf{0 \% Stragglers}}                                                                       & \multicolumn{3}{c|}{\textbf{10 \% Stragglers}}                                 & \multicolumn{3}{c}{\textbf{20 \% Stragglers}}             \\ \hline
\textbf{$\alpha$}        & \textbf{party \%}        & \textbf{Random}   & \textbf{FLIPS} & \textbf{OORT}     & \textbf{GradCls}  & \multicolumn{1}{c|}{\textbf{TiFL}}     & \textbf{FLIPS}    & \textbf{OORT}     & \multicolumn{1}{c|}{\textbf{TiFL}}     & \textbf{FLIPS}    & \textbf{OORT}     & \textbf{TiFL}     \\ \hline
0.3                   & 20                             & \textgreater{}400 & 157            & 373               & \textgreater{}400 & \multicolumn{1}{c|}{\textgreater{}400} & 193               & \textgreater{}400 & \multicolumn{1}{c|}{\textgreater{}400} & 192               & \textgreater{}400 & \textgreater{}400 \\
0.3                   & 15                             & \textgreater{}400 & 172            & 187               & \textgreater{}400 & \multicolumn{1}{c|}{\textgreater{}400} & 263               & \textgreater{}400 & \multicolumn{1}{c|}{\textgreater{}400} & 214               & \textgreater{}400 & \textgreater{}400 \\
0.6                   & 20                             & \textgreater{}400 & 242            & 280               & \textgreater{}400 & \multicolumn{1}{c|}{\textgreater{}400} & \textgreater{}400 & \textgreater{}400 & \multicolumn{1}{c|}{\textgreater{}400} & 326               & \textgreater{}400 & \textgreater{}400 \\
\end{tabular}

\label{mit:rounds}
\end{table*}

\begin{table*}[]
\footnotesize
\centering
\caption{MIT ECG Dataset: highest accuracy attained within the rounds threshold}
\begin{tabular}{cc|ccccccccccc}
\hline
\multicolumn{2}{c|}{\multirow{2}{*}{\textbf{Setting}}} & \multicolumn{11}{c}{\textbf{Highest accuracy attained, FL Algorithm : FedYoGi}}                                                                                                                                                                         \\ \cline{3-13} 
\multicolumn{2}{c|}{}                                  & \multicolumn{5}{c|}{\textbf{0 \% Stragglers}}                                                            & \multicolumn{3}{c|}{\textbf{10 \% Stragglers}}                      & \multicolumn{3}{c}{\textbf{20 \% Stragglers}}  \\ \hline
\textbf{$\alpha$}        & \textbf{party \%}        & \textbf{Random} & \textbf{FLIPS} & \textbf{OORT} & \textbf{GradCls} & \multicolumn{1}{c|}{\textbf{TiFL}} & \textbf{FLIPS} & \textbf{OORT} & \multicolumn{1}{c|}{\textbf{TiFL}} & \textbf{FLIPS} & \textbf{OORT} & \textbf{TiFL} \\ \hline
0.3                   & 20                             & 44.86           & 78.53          & 61.75         & 48.62            & \multicolumn{1}{c|}{48.21}         & 74.66          & 37.27         & \multicolumn{1}{c|}{50.51}         & 74.20          & 43.64         & 44.37         \\
0.3                   & 15                             & 45.48           & 76.92          & 71.35         & 45.40            & \multicolumn{1}{c|}{47.67}         & 71.09          & 49.19         & \multicolumn{1}{c|}{48.41}         & 67.14          & 42.34         & 49.09         \\
0.6                   & 20                             & 48.55           & 63.79          & 63.74         & 48.10            & \multicolumn{1}{c|}{41.97}         & 57.21          & 42.55         & \multicolumn{1}{c|}{52.18}         & 60.50          & 47.15         & 48.20         \\
\end{tabular}

\label{mit:accuracy}
\end{table*}

\begin{table*}[]
\footnotesize\centering
\caption{HAM10000 (Skin lesion) dataset : Rounds required to attain Target Accuracy (60\%)}
\begin{tabular}{cc|ccccccccccc}
\hline
\multicolumn{2}{c|}{\multirow{2}{*}{\textbf{Setting}}} & \multicolumn{11}{c}{\textbf{Rounds required to attain 60 \% accuracy, FL Algorithm : FedYoGi}}                                                                                                                                                                \\ \cline{3-13} 
\multicolumn{2}{c|}{}                                  & \multicolumn{5}{c|}{\textbf{0 \% Stragglers}}                                                                       & \multicolumn{3}{c|}{\textbf{10 \% Stragglers}}                              & \multicolumn{3}{c}{\textbf{20 \% Stragglers}}          \\ \hline
\textbf{$\alpha$}        & \textbf{party \%}        & \textbf{Random}   & \textbf{FLIPS} & \textbf{OORT}     & \textbf{GradCls}  & \multicolumn{1}{c|}{\textbf{TiFL}}     & \textbf{FLIPS} & \textbf{OORT}     & \multicolumn{1}{c|}{\textbf{TiFL}}     & \textbf{FLIPS} & \textbf{OORT}     & \textbf{TiFL}     \\ \hline
0.3                   & 20                             & \textgreater{}400 & 167            & 262               & \textgreater{}400 & \multicolumn{1}{c|}{\textgreater{}400} & 211            & \textgreater{}400 & \multicolumn{1}{c|}{\textgreater{}400} & 202            & \textgreater{}400 & \textgreater{}400 \\
0.3                   & 15                             & \textgreater{}400 & 190            & \textgreater{}400 & \textgreater{}400 & \multicolumn{1}{c|}{\textgreater{}400} & 263            & \textgreater{}400 & \multicolumn{1}{c|}{\textgreater{}400} & 190            & \textgreater{}400 & \textgreater{}400 \\
0.6                   & 20                             & \textgreater{}400 & 231            & 306               & \textgreater{}400 & \multicolumn{1}{c|}{\textgreater{}400} & 313            & \textgreater{}400 & \multicolumn{1}{c|}{\textgreater{}400} & 388            & \textgreater{}400 & \textgreater{}400 \\
\end{tabular}
\label{sk:rounds}
\end{table*}

\begin{table*}[]
\footnotesize\centering
\caption{HAM10000 (Skin lesion) dataset : highest accuracy attained in the rounds threshold}
\begin{tabular}{cc|ccccccccccc}
\hline
\multicolumn{2}{c|}{\multirow{2}{*}{\textbf{Setting}}} & \multicolumn{11}{c}{\textbf{Highest accuracy attained, FL Algorithm : FedYoGi}}                                                                                                                                                                         \\ \cline{3-13} 
\multicolumn{2}{c|}{}                                  & \multicolumn{5}{c|}{\textbf{0 \% Stragglers}}                                                            & \multicolumn{3}{c|}{\textbf{10 \% Stragglers}}                      & \multicolumn{3}{c}{\textbf{20 \% Stragglers}}  \\ \hline
\textbf{$\alpha$}        & \textbf{party \%}        & \textbf{Random} & \textbf{FLIPS} & \textbf{OORT} & \textbf{GradCls} & \multicolumn{1}{c|}{\textbf{TiFL}} & \textbf{FLIPS} & \textbf{OORT} & \multicolumn{1}{c|}{\textbf{TiFL}} & \textbf{FLIPS} & \textbf{OORT} & \textbf{TiFL} \\ \hline
0.3                   & 20                             & 48.26           & 66.76          & 61.12         & 46.48            & \multicolumn{1}{c|}{42.39}         & 63.39          & 46.28         & \multicolumn{1}{c|}{49.11}         & 64.13          & 38.25         & 41.59         \\
0.3                   & 15                             & 41.35           & 66.41          & 59.86         & 45.25            & \multicolumn{1}{c|}{43.70}         & 62.76          & 43.14         & \multicolumn{1}{c|}{47.09}         & 64.42          & 49.86         & 51.81         \\
0.6                   & 20                             & 46.50           & 62.84          & 62.36         & 54.74            & \multicolumn{1}{c|}{45.17}         & 60.58          & 41.94         & \multicolumn{1}{c|}{44.72}         & 60.71          & 43.08         & 44.81         \\
\end{tabular}

\label{sk:accuracy}
\end{table*}

\begin{table*}[]
\footnotesize\centering
\caption{FEMNIST : Rounds required to attain Target Accuracy (80\%)}
\begin{tabular}{cc|ccccccccccc}
\hline
\multicolumn{2}{c|}{\multirow{2}{*}{\textbf{Setting}}} & \multicolumn{11}{c}{\textbf{Rounds required to attain 80 \% accuracy, FL Algorithm : FedYoGi}}                                                                                                                                     \\ \cline{3-13} 
\multicolumn{2}{c|}{}                                  & \multicolumn{5}{c|}{\textbf{0 \% Stragglers}}                                                            & \multicolumn{3}{c|}{\textbf{10 \% Stragglers}}                      & \multicolumn{3}{c}{\textbf{20 \% Stragglers}}  \\ \hline
\textbf{$\alpha$}        & \textbf{party \%}        & \textbf{Random} & \textbf{FLIPS} & \textbf{OORT} & \textbf{GradCls} & \multicolumn{1}{c|}{\textbf{TiFL}} & \textbf{FLIPS} & \textbf{OORT} & \multicolumn{1}{c|}{\textbf{TiFL}} & \textbf{FLIPS} & \textbf{OORT} & \textbf{TiFL} \\ \hline
0.3                   & 20                             & 146             & 62             & 81            & 168              & \multicolumn{1}{c|}{124}           & 69             & 102           & \multicolumn{1}{c|}{113}           & 78             & 85            & 194           \\
0.3                   & 15                             & 181             & 62             & 83            & 168              & \multicolumn{1}{c|}{127}           & 76             & 105           & \multicolumn{1}{c|}{141}           & 79             & 103           & 106           \\
0.6                   & 20                             & 107             & 68             & 66            & 89               & \multicolumn{1}{c|}{104}           & 89             & 69            & \multicolumn{1}{c|}{103}           & 89             & 84            & 94            \\
\end{tabular}
\label{femnist:yogi:rounds}
\end{table*}

\begin{table*}[]
\footnotesize\centering
\caption{FEMNIST dataset : highest accuracy attained in the rounds threshold}

\begin{tabular}{cc|ccccccccccc}
\hline
\multicolumn{2}{c|}{\multirow{2}{*}{\textbf{Setting}}} & \multicolumn{11}{c}{\textbf{Highest accuracy attained, FL Algorithm : FedYoGi}}                                                                                                                                                                         \\ \cline{3-13} 
\multicolumn{2}{c|}{}                                  & \multicolumn{5}{c|}{\textbf{0 \% Stragglers}}                                                            & \multicolumn{3}{c|}{\textbf{10 \% Stragglers}}                      & \multicolumn{3}{c}{\textbf{20 \% Stragglers}}  \\ \hline
\textbf{$\alpha$}        & \textbf{party \%}        & \textbf{Random} & \textbf{FLIPS} & \textbf{OORT} & \textbf{GradCls} & \multicolumn{1}{c|}{\textbf{TiFL}} & \textbf{FLIPS} & \textbf{OORT} & \multicolumn{1}{c|}{\textbf{TiFL}} & \textbf{FLIPS} & \textbf{OORT} & \textbf{TiFL} \\ \hline
0.3                   & 20                             & 80.97           & 86.60          & 85.27         & 80.97            & \multicolumn{1}{c|}{82.17}         & 86.75          & 83.21         & \multicolumn{1}{c|}{81.95}         & 85.85          & 84.69         & 80.20         \\
0.3                   & 15                             & 82.60           & 86.86          & 84.61         & 82.51            & \multicolumn{1}{c|}{81.44}         & 86.68          & 85.24         & \multicolumn{1}{c|}{80.36}         & 86.78          & 84.74         & 80.89         \\
0.6                   & 20                             & 83.94           & 85.37          & 85.36         & 84.21            & \multicolumn{1}{c|}{84.23}         & 85.13          & 85.44         & \multicolumn{1}{c|}{84.39}         & 85.69          & 85.00         & 84.35         \\
\end{tabular}
\label{femnist:yogi:acc}

\end{table*}

\begin{table*}[]
\footnotesize\centering
\caption{Fashion MNIST : Rounds required to attain Target Accuracy (80\%)}
\begin{tabular}{cc|ccccccccccc}
\hline
\multicolumn{2}{c|}{\multirow{2}{*}{\textbf{Setting}}} & \multicolumn{11}{c}{\textbf{Rounds required to attain 80 \% accuracy, FL Algorithm : FedYoGi}}                                                                                                                                     \\ \cline{3-13} 
\multicolumn{2}{c|}{}                                  & \multicolumn{5}{c|}{\textbf{0 \% Stragglers}}                                                            & \multicolumn{3}{c|}{\textbf{10 \% Stragglers}}                      & \multicolumn{3}{c}{\textbf{20 \% Stragglers}}  \\ \hline
\textbf{$\alpha$}        & \textbf{party \%}        & \textbf{Random} & \textbf{FLIPS} & \textbf{OORT} & \textbf{GradCls} & \multicolumn{1}{c|}{\textbf{TiFL}} & \textbf{FLIPS} & \textbf{OORT} & \multicolumn{1}{c|}{\textbf{TiFL}} & \textbf{FLIPS} & \textbf{OORT} & \textbf{TiFL} \\ \hline
0.3                   & 20                             & 62              & 48             & 72            & 62               & \multicolumn{1}{c|}{92}            & 44             & 83            & \multicolumn{1}{c|}{101}           & 48             & 72            & 91            \\
0.3                   & 15                             & 60              & 51             & 74            & 58               & \multicolumn{1}{c|}{107}           & 53             & 69            & \multicolumn{1}{c|}{91}            & 42             & 82            & 104           \\
0.6                   & 20                             & 52              & 42             & 62            & 51               & \multicolumn{1}{c|}{73}            & 37             & 65            & \multicolumn{1}{c|}{71}            & 48             & 63            & 81            \\
\end{tabular}
\label{fashion:yogi:rounds}
\end{table*}

\begin{table*}[]
\footnotesize\centering
\caption{Fashion MNIST dataset : highest accuracy attained in the rounds threshold}
\begin{tabular}{cc|ccccccccccc}
\hline
\multicolumn{2}{c|}{\multirow{2}{*}{\textbf{Setting}}} & \multicolumn{11}{c}{\textbf{Highest accuracy attained, FL Algorithm : FedYoGi}}                                                                                                                                                                         \\ \cline{3-13} 
\multicolumn{2}{c|}{}                                  & \multicolumn{5}{c|}{\textbf{0 \% Stragglers}}                                                            & \multicolumn{3}{c|}{\textbf{10 \% Stragglers}}                      & \multicolumn{3}{c}{\textbf{20 \% Stragglers}}  \\ \hline
\textbf{$\alpha$}        & \textbf{party \%}        & \textbf{Random} & \textbf{FLIPS} & \textbf{OORT} & \textbf{GradCls} & \multicolumn{1}{c|}{\textbf{TiFL}} & \textbf{FLIPS} & \textbf{OORT} & \multicolumn{1}{c|}{\textbf{TiFL}} & \textbf{FLIPS} & \textbf{OORT} & \textbf{TiFL} \\ \hline
0.3                   & 20                             & 83.92           & 85.14          & 82.45         & 83.88            & \multicolumn{1}{c|}{81.93}         & 85.29          & 82.24         & \multicolumn{1}{c|}{81.82}         & 84.51          & 83.16         & 81.99         \\
0.3                   & 15                             & 83.62           & 84.75          & 82.44         & 83.91            & \multicolumn{1}{c|}{81.85}         & 84.98          & 82.35         & \multicolumn{1}{c|}{82.53}         & 85.02          & 82.48         & 82.19         \\
0.6                   & 20                             & 84.49           & 85.56          & 83.12         & 84.65            & \multicolumn{1}{c|}{83.29}         & 85.70          & 82.89         & \multicolumn{1}{c|}{82.84}         & 85.03          & 83.18         & 82.56         \\
\end{tabular}
\label{fashion:yogi:accuracy}
\end{table*}

\begin{table*}[]
\footnotesize
\centering
\caption{MIT ECG Dataset : Rounds required to attain Target Accuracy (60\%)}

\begin{tabular}{cc|ccccccccccc}
\hline
\multicolumn{2}{c|}{\multirow{2}{*}{\textbf{Setting}}} & \multicolumn{11}{c}{\textbf{Rounds required to attain 60 \% accuracy, FL Algorithm : FedProx}}                                                                                                                                                                \\ \cline{3-13} 
\multicolumn{2}{c|}{}                                  & \multicolumn{5}{c|}{\textbf{0 \% Stragglers}}                                                                       & \multicolumn{3}{c|}{\textbf{10 \% Stragglers}}                              & \multicolumn{3}{c}{\textbf{20 \% Stragglers}}             \\ \hline
\textbf{$\alpha$}        & \textbf{party \%}        & \textbf{Random}   & \textbf{FLIPS} & \textbf{OORT}     & \textbf{GradCls}  & \multicolumn{1}{c|}{\textbf{TiFL}}     & \textbf{FLIPS} & \textbf{OORT}     & \multicolumn{1}{c|}{\textbf{TiFL}}     & \textbf{FLIPS}    & \textbf{OORT}     & \textbf{TiFL}     \\ \hline
0.3                   & 20                             & \textgreater{}400 & 129            & 198               & \textgreater{}400 & \multicolumn{1}{c|}{\textgreater{}400} & 143            & \textgreater{}400 & \multicolumn{1}{c|}{\textgreater{}400} & 255               & \textgreater{}400 & \textgreater{}400 \\
0.3                   & 15                             & \textgreater{}400 & 146            & 197               & \textgreater{}400 & \multicolumn{1}{c|}{\textgreater{}400} & 204            & \textgreater{}400 & \multicolumn{1}{c|}{\textgreater{}400} & 215               & \textgreater{}400 & \textgreater{}400 \\
0.6                   & 20                             & \textgreater{}400 & 182            & 334               & \textgreater{}400 & \multicolumn{1}{c|}{\textgreater{}400} & 383            & \textgreater{}400 & \multicolumn{1}{c|}{\textgreater{}400} & 389               & \textgreater{}400 & \textgreater{}400 \\
\end{tabular}
\label{ecg:rounds:prox}
\end{table*}

\begin{table*}[]
\footnotesize
\centering
\caption{MIT ECG Dataset: highest accuracy attained within the rounds threshold}
\begin{tabular}{cc|ccccccccccc}
\hline
\multicolumn{2}{c|}{\multirow{2}{*}{\textbf{Setting}}} & \multicolumn{11}{c}{\textbf{Highest accuracy attained, FL Algorithm : FedProx}}                                                                                                                                                                         \\ \cline{3-13} 
\multicolumn{2}{c|}{}                                  & \multicolumn{5}{c|}{\textbf{0 \% Stragglers}}                                                            & \multicolumn{3}{c|}{\textbf{10 \% Stragglers}}                      & \multicolumn{3}{c}{\textbf{20 \% Stragglers}}  \\ \hline
\textbf{$\alpha$}        & \textbf{party \%}        & \textbf{Random} & \textbf{FLIPS} & \textbf{OORT} & \textbf{GradCls} & \multicolumn{1}{c|}{\textbf{TiFL}} & \textbf{FLIPS} & \textbf{OORT} & \multicolumn{1}{c|}{\textbf{TiFL}} & \textbf{FLIPS} & \textbf{OORT} & \textbf{TiFL} \\ \hline
0.3                   & 20                             & 46.39           & 76.25          & 72.31         & 48.99            & \multicolumn{1}{c|}{41.81}         & 75.26          & 46.94         & \multicolumn{1}{c|}{50.09}         & 68.14          & 46.40         & 44.64         \\
0.3                   & 15                             & 50.63           & 74.82          & 71.29         & 46.58            & \multicolumn{1}{c|}{50.40}         & 72.48          & 45.03         & \multicolumn{1}{c|}{51.09}         & 70.24          & 46.22         & 46.75         \\
0.6                   & 20                             & 45.18           & 65.58          & 61.40         & 44.86            & \multicolumn{1}{c|}{53.83}         & 60.16          & 46.04         & \multicolumn{1}{c|}{55.04}         & 60.10          & 49.87         & 54.86         \\
\end{tabular}

\label{ecg:accuracy:prox}
\end{table*}

\begin{table*}[]
\footnotesize
\centering
\caption{HAM10000 (Skin Lesion) dataset : Rounds required to attain Target Accuracy (60\%)}
\begin{tabular}{cc|ccccccccccc}
\hline
\multicolumn{2}{c|}{\multirow{2}{*}{\textbf{Setting}}} & \multicolumn{11}{c}{\textbf{Rounds required to attain 60 \% accuracy, FL Algorithm : FedProx}}                                                                                                                                                                \\ \cline{3-13} 
\multicolumn{2}{c|}{}                                  & \multicolumn{5}{c|}{\textbf{0 \% Stragglers}}                                                                       & \multicolumn{3}{c|}{\textbf{10 \% Stragglers}}                                 & \multicolumn{3}{c}{\textbf{20 \% Stragglers}}          \\ \hline
\textbf{$\alpha$}        & \textbf{party \%}        & \textbf{Random}   & \textbf{FLIPS} & \textbf{OORT}     & \textbf{GradCls}  & \multicolumn{1}{c|}{\textbf{TiFL}}     & \textbf{FLIPS}    & \textbf{OORT}     & \multicolumn{1}{c|}{\textbf{TiFL}}     & \textbf{FLIPS} & \textbf{OORT}     & \textbf{TiFL}     \\ \hline
0.3                   & 20                             & \textgreater{}400 & 151            & 323               & \textgreater{}400 & \multicolumn{1}{c|}{\textgreater{}400} & 206               & \textgreater{}400 & \multicolumn{1}{c|}{\textgreater{}400} & 172            & \textgreater{}400 & \textgreater{}400 \\
0.3                   & 15                             & \textgreater{}400 & 201            & 298               & \textgreater{}400 & \multicolumn{1}{c|}{\textgreater{}400} & 198               & \textgreater{}400 & \multicolumn{1}{c|}{\textgreater{}400} & 198            & \textgreater{}400 & \textgreater{}400 \\
0.6                   & 20                             & \textgreater{}400 & 276            & \textgreater{}400 & \textgreater{}400 & \multicolumn{1}{c|}{\textgreater{}400} & \textgreater{}400 & \textgreater{}400 & \multicolumn{1}{c|}{\textgreater{}400} & 364            & \textgreater{}400 & \textgreater{}400 \\
\end{tabular}
\label{sk:rounds:prox}
\end{table*}

\begin{table*}[]
\footnotesize
\centering
\caption{HAM10000 (Skin Lesion) dataset: highest accuracy attained within the rounds threshold}

\begin{tabular}{cc|ccccccccccc}
\hline
\multicolumn{2}{c|}{\multirow{2}{*}{\textbf{Setting}}} & \multicolumn{11}{c}{\textbf{Highest accuracy attained, FL Algorithm : FedProx}}                                                                                                                                                                         \\ \cline{3-13} 
\multicolumn{2}{c|}{}                                  & \multicolumn{5}{c|}{\textbf{0 \% Stragglers}}                                                            & \multicolumn{3}{c|}{\textbf{10 \% Stragglers}}                      & \multicolumn{3}{c}{\textbf{20 \% Stragglers}}  \\ \hline
\textbf{$\alpha$}        & \textbf{party \%}        & \textbf{Random} & \textbf{FLIPS} & \textbf{OORT} & \textbf{GradCls} & \multicolumn{1}{c|}{\textbf{TiFL}} & \textbf{FLIPS} & \textbf{OORT} & \multicolumn{1}{c|}{\textbf{TiFL}} & \textbf{FLIPS} & \textbf{OORT} & \textbf{TiFL} \\ \hline
0.3                   & 20                             & 47.08           & 64.53          & 60.32         & 46.84            & \multicolumn{1}{c|}{}              & 65.76          & 42.13         & \multicolumn{1}{c|}{46.07}         & 67.15          & 48.24         & 51.71         \\
0.3                   & 15                             & 41.59           & 66.71          & 62.25         & 46.16            & \multicolumn{1}{c|}{46.48}         & 65.55          & 44.07         & \multicolumn{1}{c|}{40.26}         & 66.74          & 43.23         & 39.01         \\
0.6                   & 20                             & 43.66           & 63.55          & 58.67         & 53.65            & \multicolumn{1}{c|}{54.4}          & 58.89          & 50.15         & \multicolumn{1}{c|}{54.36}         & 60.87          & 51.07         & 46.38         \\
\end{tabular}

\label{sk:accuracy:prox}

\end{table*}

\begin{table*}[]
\footnotesize
\centering
\caption{FEMNIST dataset : Rounds required to attain Target Accuracy (80\%)}
\begin{tabular}{cc|ccccccccccc}
\hline
\multicolumn{2}{c|}{\multirow{2}{*}{\textbf{Setting}}} & \multicolumn{11}{c}{\textbf{Rounds required to attain 80 \% accuracy, Optimizer : FedProx}}                                                                                                                                     \\ \cline{3-13} 
\multicolumn{2}{c|}{}                                  & \multicolumn{5}{c|}{\textbf{0 \% Stragglers}}                                                            & \multicolumn{3}{c|}{\textbf{10 \% Stragglers}}                     & \multicolumn{3}{c}{\textbf{20 \% Stragglers}}  \\ \hline
\textbf{$\alpha$}        & \textbf{party \%}        & \textbf{Random} & \textbf{FLIPS} & \textbf{OORT} & \textbf{GradCls} & \multicolumn{1}{c|}{\textbf{TiFL}} & \textbf{FLIPS} & \textbf{OORT} & \multicolumn{1}{c|}{\textbf{TiFL}} & \textbf{FLIPS} & \textbf{OORT} & \textbf{TiFL} \\ \hline
0.3                   & 20                             & 128             & 47             & 71            & 157              & \multicolumn{1}{c|}{103}           & 65             & 130           & \multicolumn{1}{c|}{98}            & 78             & 128           & 146           \\
0.3                   & 15                             & 104             & 54             & 70            & 149              & \multicolumn{1}{c|}{111}           & 72             & 118           & \multicolumn{1}{c|}{110}           & 67             & 156           & 116           \\
0.6                   & 20                             & 84              & 62             & 53            & 84               & \multicolumn{1}{c|}{110}           & 90             & 82            & \multicolumn{1}{c|}{108}           & 80             & 77            & 98            \\
\end{tabular}
\end{table*}

\begin{table*}[]
\footnotesize
\caption{FEMNIST dataset: highest accuracy attained within the rounds threshold}
\begin{tabular}{cc|ccccccccccc}
\hline
\multicolumn{2}{c|}{\multirow{2}{*}{\textbf{Setting}}} & \multicolumn{11}{c}{\textbf{Highest accuracy attained, FL Algorithm : FedProx}}                                                                                                                                                                         \\ \cline{3-13} 
\multicolumn{2}{c|}{}                                  & \multicolumn{5}{c|}{\textbf{0 \% Stragglers}}                                                            & \multicolumn{3}{c|}{\textbf{10 \% Stragglers}}                      & \multicolumn{3}{c}{\textbf{20 \% Stragglers}}  \\ \hline
\textbf{$\alpha$}        & \textbf{party \%}        & \textbf{Random} & \textbf{FLIPS} & \textbf{OORT} & \textbf{GradCls} & \multicolumn{1}{c|}{\textbf{TiFL}} & \textbf{FLIPS} & \textbf{OORT} & \multicolumn{1}{c|}{\textbf{TiFL}} & \textbf{FLIPS} & \textbf{OORT} & \textbf{TiFL} \\ \hline
0.3                   & 20                             & 82.80           & 90.43          & 86.59         & 81.78            & \multicolumn{1}{c|}{82.96}         & 87.33          & 83.21         & \multicolumn{1}{c|}{82.53}         & 86.56          & 83.81         & 80.47         \\
0.3                   & 15                             & 83.24           & 89.72          & 86.51         & 83.34            & \multicolumn{1}{c|}{83.33}         & 87.02          & 83.81         & \multicolumn{1}{c|}{82.73}         & 86.82          & 82.43         & 81.97         \\
0.6                   & 20                             & 85.60           & 88.99          & 87.45         & 85.05            & \multicolumn{1}{c|}{85.29}         & 85.23          & 86.11         & \multicolumn{1}{c|}{84.29}         & 85.72          & 85.91         & 84.87         \\
\end{tabular}
\end{table*}

\begin{table*}[]
\footnotesize
\centering
\caption{Fashion MNIST Dataset: : Rounds required to attain Target Accuracy (80\%)}

\begin{tabular}{cc|ccccccccccc}
\hline
\multicolumn{2}{c|}{\multirow{2}{*}{\textbf{Setting}}} & \multicolumn{11}{c}{\textbf{Rounds required to attain 80 \% accuracy, Optimizer : FedProx}}                                                                                                                                     \\ \cline{3-13} 
\multicolumn{2}{c|}{}                                  & \multicolumn{5}{c|}{\textbf{0 \% Stragglers}}                                                            & \multicolumn{3}{c|}{\textbf{10 \% Stragglers}}                      & \multicolumn{3}{c}{\textbf{20 \% Stragglers}}  \\ \hline
\textbf{$\alpha$}        & \textbf{party \%}        & \textbf{Random} & \textbf{FLIPS} & \textbf{OORT} & \textbf{GradCls} & \multicolumn{1}{c|}{\textbf{TiFL}} & \textbf{FLIPS} & \textbf{OORT} & \multicolumn{1}{c|}{\textbf{TiFL}} & \textbf{FLIPS} & \textbf{OORT} & \textbf{TiFL} \\ \hline
0.3                   & 20                             & 74              & 47             & 83            & 66               & \multicolumn{1}{c|}{82}            & 48             & 74            & \multicolumn{1}{c|}{101}           & 45             & 70            & 91            \\
0.3                   & 15                             & 62              & 42             & 75            & 69               & \multicolumn{1}{c|}{78}            & 48             & 82            & \multicolumn{1}{c|}{91}            & 48             & 71            & 104           \\
0.6                   & 20                             & 52              & 46             & 64            & 49               & \multicolumn{1}{c|}{70}            & 36             & 70            & \multicolumn{1}{c|}{71}            & 47             & 71            & 82            \\
\end{tabular}
\label{fashion:prox:rounds}
\end{table*}

\begin{table*}[]
\footnotesize
\centering
\caption{Fashion MNIST Dataset: : highest accuracy attained within the rounds threshold}

\begin{tabular}{cc|ccccccccccc}
\hline
\multicolumn{2}{c|}{\multirow{2}{*}{\textbf{Setting}}} & \multicolumn{11}{c}{\textbf{Highest accuracy attained, FL Algorithm : FedProx}}                                                                                                                                                                         \\ \cline{3-13} 
\multicolumn{2}{c|}{}                                  & \multicolumn{5}{c|}{\textbf{0 \% Stragglers}}                                                            & \multicolumn{3}{c|}{\textbf{10 \% Stragglers}}                      & \multicolumn{3}{c}{\textbf{20 \% Stragglers}}  \\ \hline
\textbf{$\alpha$}        & \textbf{party \%}        & \textbf{Random} & \textbf{FLIPS} & \textbf{OORT} & \textbf{GradCls} & \multicolumn{1}{c|}{\textbf{TiFL}} & \textbf{FLIPS} & \textbf{OORT} & \multicolumn{1}{c|}{\textbf{TiFL}} & \textbf{FLIPS} & \textbf{OORT} & \textbf{TiFL} \\ \hline
0.3                   & 20                             & 83.91           & 85.04          & 82.52         & 83.46            & \multicolumn{1}{c|}{82.36}         & 85.10          & 82.52         & \multicolumn{1}{c|}{81.82}         & 85.31          & 82.89         & 81.99         \\
0.3                   & 15                             & 83.66           & 85.46          & 82.61         & 83.83            & \multicolumn{1}{c|}{82.01}         & 84.98          & 81.97         & \multicolumn{1}{c|}{82.53}         & 84.93          & 82.16         & 82.19         \\
0.6                   & 20                             & 84.48           & 85.52          & 82.86         & 84.67            & \multicolumn{1}{c|}{83.11}         & 86.14          & 82.60         & \multicolumn{1}{c|}{82.84}         & 85.02          & 82.95         & 82.56         \\
\end{tabular}
\label{fashion:prox:accuracy}
\end{table*}

\begin{table*}[]
\footnotesize
\centering
\caption{MIT ECG Dataset: Rounds required to attain Target Accuracy (60\%)}
\begin{tabular}{cc|ccccccccccc}
\hline
\multicolumn{2}{c|}{\multirow{2}{*}{\textbf{Setting}}} & \multicolumn{11}{c}{\textbf{Rounds required to attain 60 \% accuracy, FL Algorithm : FedAvg}}                                                                                                                                                                    \\ \cline{3-13} 
\multicolumn{2}{c|}{}                                  & \multicolumn{5}{c|}{\textbf{0 \% Stragglers}}                                                                       & \multicolumn{3}{c|}{\textbf{10 \% Stragglers}}                                 & \multicolumn{3}{c}{\textbf{20 \% Stragglers}}             \\ \hline
\textbf{$\alpha$}        & \textbf{party \%}        & \textbf{Random}   & \textbf{FLIPS} & \textbf{OORT}     & \textbf{GradCls}  & \multicolumn{1}{c|}{\textbf{TiFL}}     & \textbf{FLIPS}    & \textbf{OORT}     & \multicolumn{1}{c|}{\textbf{TiFL}}     & \textbf{FLIPS}    & \textbf{OORT}     & \textbf{TiFL}     \\ \hline
0.3                   & 20                             & \textgreater{}400 & 136            & 344               & \textgreater{}400 & \multicolumn{1}{c|}{\textgreater{}400} & 210               & \textgreater{}400 & \multicolumn{1}{c|}{\textgreater{}400} & 200               & \textgreater{}400 & \textgreater{}400 \\
0.3                   & 15                             & \textgreater{}400 & 162            & 192               & \textgreater{}400 & \multicolumn{1}{c|}{\textgreater{}400} & 263               & \textgreater{}400 & \multicolumn{1}{c|}{\textgreater{}400} & 214               & \textgreater{}400 & \textgreater{}400 \\
0.6                   & 20                             & \textgreater{}400 & 378            & 393               & \textgreater{}400 & \multicolumn{1}{c|}{\textgreater{}400} & \textgreater{}400 & \textgreater{}400 & \multicolumn{1}{c|}{\textgreater{}400} & 397               & \textgreater{}400 & \textgreater{}400 \\
\end{tabular}

\label{ecg:rounds:fedavg}

\end{table*}

\begin{table*}[]
\footnotesize
\centering
\caption{MIT BIH ECG Dataset: : highest accuracy attained within the rounds threshold}
\begin{tabular}{cc|ccccccccccc}
\hline
\multicolumn{2}{c|}{\multirow{2}{*}{\textbf{Setting}}} & \multicolumn{11}{c}{\textbf{Highest accuracy attained, FL Algorithm : FedAvg}}                                                                                                                                                                         \\ \cline{3-13} 
\multicolumn{2}{c|}{}                                  & \multicolumn{5}{c|}{\textbf{0 \% Stragglers}}                                                            & \multicolumn{3}{c|}{\textbf{10 \% Stragglers}}                      & \multicolumn{3}{c}{\textbf{20 \% Stragglers}}  \\ \hline
\textbf{$\alpha$}        & \textbf{party \%}        & \textbf{Random} & \textbf{FLIPS} & \textbf{OORT} & \textbf{GradCls} & \multicolumn{1}{c|}{\textbf{TiFL}} & \textbf{FLIPS} & \textbf{OORT} & \multicolumn{1}{c|}{\textbf{TiFL}} & \textbf{FLIPS} & \textbf{OORT} & \textbf{TiFL} \\ \hline
0.3                   & 20                             & 47.92           & 73.33          & 63.02         & 45.26            & \multicolumn{1}{c|}{45.76}         & 73.16          & 36.53         & \multicolumn{1}{c|}{48.53}         & 72.71          & 42.77         & 46.48         \\
0.3                   & 15                             & 48.06           & 72.81          & 70.12         & 44.09            & \multicolumn{1}{c|}{48.16}         & 69.67          & 48.21         & \multicolumn{1}{c|}{49.32}         & 65.80          & 41.49         & 53.75         \\
0.6                   & 20                             & 51.97           & 63.76          & 60.67         & 48.10            & \multicolumn{1}{c|}{46.86}         & 56.07          & 41.70         & \multicolumn{1}{c|}{44.86}         & 60.29          & 46.20         & 52.05         \\
\end{tabular}
\label{ecg:accuracy:avg}
\end{table*}

\begin{table*}[]
\footnotesize
\centering
\caption{HAM10000 (Skin Lesion) Dataset: : Rounds required to attain Target Accuracy (60\%)}
\begin{tabular}{cc|ccccccccccc}
\hline
\multicolumn{2}{c|}{\multirow{2}{*}{\textbf{Setting}}} & \multicolumn{11}{c}{\textbf{Rounds required to attain 60 \% accuracy, FL Algorithm : FedAvg}}                                                                                                                                                              \\ \cline{3-13} 
\multicolumn{2}{c|}{}                                  & \multicolumn{5}{c|}{\textbf{0 \% Stragglers}}                                                                       & \multicolumn{3}{c|}{\textbf{10 \% Stragglers}}                              & \multicolumn{3}{c}{\textbf{20 \% Stragglers}}          \\ \hline
\textbf{$\alpha$}        & \textbf{party \%}        & \textbf{Random}   & \textbf{FLIPS} & \textbf{OORT}     & \textbf{GradCls}  & \multicolumn{1}{c|}{\textbf{TiFL}}     & \textbf{FLIPS} & \textbf{OORT}     & \multicolumn{1}{c|}{\textbf{TiFL}}     & \textbf{FLIPS} & \textbf{OORT}     & \textbf{TiFL}     \\ \hline
0.3                   & 20                             & \textgreater{}400 & 329            & 271               & \textgreater{}400 & \multicolumn{1}{c|}{\textgreater{}400} & 250            & \textgreater{}400 & \multicolumn{1}{c|}{\textgreater{}400} & 234            & \textgreater{}400 & \textgreater{}400 \\
0.3                   & 15                             & \textgreater{}400 & 300            & 323               & \textgreater{}400 & \multicolumn{1}{c|}{\textgreater{}400} & 201            & \textgreater{}400 & \multicolumn{1}{c|}{\textgreater{}400} & 217            & \textgreater{}400 & \textgreater{}400 \\
0.6                   & 20                             & \textgreater{}400 & 300            & 385               & \textgreater{}400 & \multicolumn{1}{c|}{\textgreater{}400} & 376            & \textgreater{}400 & \multicolumn{1}{c|}{\textgreater{}400} & 356            & \textgreater{}400 & \textgreater{}400 \\
\end{tabular}
\label{sk:rounds:fedavg}
\end{table*}

\begin{table*}[]
\footnotesize
\centering
\caption{HAM10000 (Skin Lesion) Dataset: : highest accuracy attained within the rounds threshold}

\begin{tabular}{cc|ccccccccccc}
\hline
\multicolumn{2}{c|}{\multirow{2}{*}{\textbf{Setting}}} & \multicolumn{11}{c}{\textbf{Highest accuracy attained, FL Algorithm : FedAvg}}                                                                                                                                                                         \\ \cline{3-13} 
\multicolumn{2}{c|}{}                                  & \multicolumn{5}{c|}{\textbf{0 \% Stragglers}}                                                            & \multicolumn{3}{c|}{\textbf{10 \% Stragglers}}                      & \multicolumn{3}{c}{\textbf{20 \% Stragglers}}  \\ \hline
\textbf{$\alpha$}        & \textbf{party \%}        & \textbf{Random} & \textbf{FLIPS} & \textbf{OORT} & \textbf{GradCls} & \multicolumn{1}{c|}{\textbf{TiFL}} & \textbf{FLIPS} & \textbf{OORT} & \multicolumn{1}{c|}{\textbf{TiFL}} & \textbf{FLIPS} & \textbf{OORT} & \textbf{TiFL} \\ \hline
0.3                   & 20                             & 46.76           & 64.79          & 62.05         & 47.56            & \multicolumn{1}{c|}{44.58}         & 62.96          & 42.99         & \multicolumn{1}{c|}{45.73}         & 63.51          & 49.50         & 51.46         \\
0.3                   & 15                             & 41.83           & 64.82          & 61.70         & 46.87            & \multicolumn{1}{c|}{45.62}         & 63.65          & 54.70         & \multicolumn{1}{c|}{56.88}         & 65.71          & 48.96         & 49.37         \\
0.6                   & 20                             & 46.50           & 62.42          & 60.56         & 54.48            & \multicolumn{1}{c|}{50}            & 60.14          & 52.50         & \multicolumn{1}{c|}{55.22}         & 60.58          & 55.48         & 57.93         \\
\end{tabular}

\label{sk:accuracy:fedavg}
\end{table*}

\begin{table*}[]
\footnotesize
\centering
\caption{FEMNIST Dataset: : Rounds required to attain Target Accuracy (80\%)}
\begin{tabular}{cc|ccccccccccc}
\hline
\multicolumn{2}{c|}{\multirow{2}{*}{\textbf{Setting}}} & \multicolumn{11}{c}{\textbf{Rounds required to attain 80 \% accuracy, Optimizer : FedAvg}}                                                                                                                                          \\ \cline{3-13} 
\multicolumn{2}{c|}{}                                  & \multicolumn{5}{c|}{\textbf{0 \% Stragglers}}                                                            & \multicolumn{3}{c|}{\textbf{10 \% Stragglers}}                          & \multicolumn{3}{c}{\textbf{20 \% Stragglers}}  \\ \hline
\textbf{$\alpha$}        & \textbf{party \%}        & \textbf{Random} & \textbf{FLIPS} & \textbf{OORT} & \textbf{GradCls} & \multicolumn{1}{c|}{\textbf{TiFL}} & \textbf{FLIPS} & \textbf{OORT} & \multicolumn{1}{c|}{\textbf{TiFL}}     & \textbf{FLIPS} & \textbf{OORT} & \textbf{TiFL} \\ \hline
0.3                   & 20                             & 130             & 46             & 71            & 168              & \multicolumn{1}{c|}{118}           & 65             & 130           & \multicolumn{1}{c|}{123}               & 78             & 128           & 153           \\
0.3                   & 15                             & 112             & 62             & 70            & 168              & \multicolumn{1}{c|}{112}           & 72             & 118           & \multicolumn{1}{c|}{\textgreater{}400} & 67             & 156           & 142           \\
0.6                   & 20                             & 99              & 69             & 53            & 89               & \multicolumn{1}{c|}{96}            & 90             & 82            & \multicolumn{1}{c|}{92}                & 80             & 77            & 102           \\
\end{tabular}
\label{femnist:rounds:fedavg}

\end{table*}

\begin{table*}[]
\footnotesize
\centering
\caption{FEMNIST Dataset: : highest accuracy attained within the rounds threshold}
\begin{tabular}{cc|ccccccccccc}
\hline
\multicolumn{2}{c|}{\multirow{2}{*}{\textbf{Setting}}} & \multicolumn{11}{c}{\textbf{Highest accuracy attained, FL Algorithm : FedAvg}}                                                                                                                                                                         \\ \cline{3-13} 
\multicolumn{2}{c|}{}                                  & \multicolumn{5}{c|}{\textbf{0 \% Stragglers}}                                                            & \multicolumn{3}{c|}{\textbf{10 \% Stragglers}}                      & \multicolumn{3}{c}{\textbf{20 \% Stragglers}}  \\ \hline
\textbf{$\alpha$}        & \textbf{party \%}        & \textbf{Random} & \textbf{FLIPS} & \textbf{OORT} & \textbf{GradCls} & \multicolumn{1}{c|}{\textbf{TiFL}} & \textbf{FLIPS} & \textbf{OORT} & \multicolumn{1}{c|}{\textbf{TiFL}} & \textbf{FLIPS} & \textbf{OORT} & \textbf{TiFL} \\ \hline
0.3                   & 20                             & 82.37           & 90.64          & 86.59         & 80.97            & \multicolumn{1}{c|}{81.78}         & 87.33          & 83.21         & \multicolumn{1}{c|}{80.09}         & 86.46          & 83.81         & 80.41         \\
0.3                   & 15                             & 82.48           & 89.08          & 86.51         & 82.51            & \multicolumn{1}{c|}{81.19}         & 87.02          & 83.81         & \multicolumn{1}{c|}{79.77}         & 86.82          & 82.43         & 80.61         \\
0.6                   & 20                             & 84.37           & 88.20          & 87.45         & 84.21            & \multicolumn{1}{c|}{85.33}         & 85.23          & 86.11         & \multicolumn{1}{c|}{85.16}         & 85.72          & 85.91         & 84.32         \\
\end{tabular}

\label{femnist:accuracy:fedavg}
\end{table*}

\begin{table*}[]
\footnotesize
\centering
\caption{Fashion MNIST Dataset: : Rounds required to attain Target Accuracy (80\%)}

\begin{tabular}{cc|ccccccccccc}
\hline
\multicolumn{2}{c|}{\multirow{2}{*}{\textbf{Setting}}} & \multicolumn{11}{c}{\textbf{Rounds required to attain 80 \% accuracy, Optimizer : FedAvg}}                                                                                                                                      \\ \cline{3-13} 
\multicolumn{2}{c|}{}                                  & \multicolumn{5}{c|}{\textbf{0 \% Stragglers}}                                                            & \multicolumn{3}{c|}{\textbf{10 \% Stragglers}}                      & \multicolumn{3}{c}{\textbf{20 \% Stragglers}}  \\ \hline
\textbf{$\alpha$}        & \textbf{party \%}        & \textbf{Random} & \textbf{FLIPS} & \textbf{OORT} & \textbf{GradCls} & \multicolumn{1}{c|}{\textbf{TiFL}} & \textbf{FLIPS} & \textbf{OORT} & \multicolumn{1}{c|}{\textbf{TiFL}} & \textbf{FLIPS} & \textbf{OORT} & \textbf{TiFL} \\ \hline
0.3                   & 20                             & 56              & 48             & 67            & 53               & \multicolumn{1}{c|}{91}            & 48             & 67            & \multicolumn{1}{c|}{100}           & 49             & 83            & 92            \\
0.3                   & 15                             & 70              & 43             & 65            & 62               & \multicolumn{1}{c|}{92}            & 52             & 80            & \multicolumn{1}{c|}{110}           & 54             & 78            & 105           \\
0.6                   & 20                             & 53              & 37             & 70            & 55               & \multicolumn{1}{c|}{74}            & 45             & 75            & \multicolumn{1}{c|}{75}            & 40             & 73            & 62            \\
\end{tabular}
\label{fashion:avg:rounds}
\end{table*}

\begin{table*}[]
\footnotesize
\centering
\caption{Fashion MNIST Dataset: : highest accuracy attained within the rounds threshold}
\begin{tabular}{cc|ccccccccccc}
\hline
\multicolumn{2}{c|}{\multirow{2}{*}{\textbf{Setting}}} & \multicolumn{11}{c}{\textbf{Highest accuracy attained, FL Algorithm : FedAvg}}                                                                                                                                                                         \\ \cline{3-13} 
\multicolumn{2}{c|}{}                                  & \multicolumn{5}{c|}{\textbf{0 \% Stragglers}}                                                            & \multicolumn{3}{c|}{\textbf{10 \% Stragglers}}                      & \multicolumn{3}{c}{\textbf{20 \% Stragglers}}  \\ \hline
\textbf{$\alpha$}        & \textbf{party \%}        & \textbf{Random} & \textbf{FLIPS} & \textbf{OORT} & \textbf{GradCls} & \multicolumn{1}{c|}{\textbf{TiFL}} & \textbf{FLIPS} & \textbf{OORT} & \multicolumn{1}{c|}{\textbf{TiFL}} & \textbf{FLIPS} & \textbf{OORT} & \textbf{TiFL} \\ \hline
0.3                   & 20                             & 84.13           & 85.00          & 82.59         & 83.99            & \multicolumn{1}{c|}{82.28}         & 84.85          & 82.73         & \multicolumn{1}{c|}{82.21}         & 85.05          & 82.39         & 81.86         \\
0.3                   & 15                             & 83.67           & 85.55          & 83.02         & 83.6             & \multicolumn{1}{c|}{81.76}         & 84.69          & 82.15         & \multicolumn{1}{c|}{81.61}         & 84.82          & 82.72         & 82.06         \\
0.6                   & 20                             & 84.95           & 85.80          & 82.82         & 84.77            & \multicolumn{1}{c|}{82.99}         & 85.53          & 82.46         & \multicolumn{1}{c|}{83.12}         & 85.23          & 82.81         & 83.35         \\
\end{tabular}
\label{fashion:avg:accuracy}
\end{table*}

\subsection{TEE Clustering Overhead}
Clustering label distributions, by itself, is fairly efficient and takes less than one second for all our datasets ($\approx$100ms for HAM10000 dataset with 200 parties and lower for other datasets on a 2.3 Ghz 4 core Intel Core i9 equipped server with 16GB RAM and 512GB SSD). The overhead of using TEEs to perform clustering is approximately 5\% (105.4 ms vs. 100.5ms) in the case of AMD SEV on a server running the aggregator. 
Hence, using TEEs gives us a reasonable way to implement private label distribution clustering for \ours.



\subsection{Data Heterogeneity}

Tables~\ref{mit:rounds} -- \ref{fashion:yogi:accuracy} 
summarize our results for FedYogi. Tables~\ref{ecg:rounds:prox} -- \ref{fashion:prox:accuracy} 
summarize our results for FedProx. Tables~\ref{ecg:rounds:fedavg} -- \ref{fashion:avg:accuracy} 
summarize our results for FedAvg. At a high level, we observe that~\ours can reach target accuracies much faster and achieve much higher peak accuracy.

When considering the number of rounds needed to reach targeted accuracies for 
the MIT-ECG (60\%) and HAM10000 (60\%) datasets,
we observe that Random selection, TiFL and Gradient Clustering take more than
400 rounds in the case of FedAvg, FedYogi and FedProx. While the performance
of Random selection is not surprising considering all the reasons outlined
so far in this paper, GradClus's performance is surprising, given that
it also clusters parties, albeit using gradients. TiFL's adaptive tiering approach is unable to group the parties with under-represented labels into a single tier, which explains its performance.
OORT's statistical utility function does enable it to perform 
better than Random, TiFL and GradClus.
This trend can also be observed from example convergence plots in 
Figures~\ref{fig:ecg:all} and \ref{fig:sk:all}. 
From Table~\ref{mit:rounds}, we observe that \ours\ reaches target accuracy
up to 1.15-1.86$\times$ faster, i.e., in fewer rounds when compared to OORT
when $\alpha~=~0.6$ and 1.08-2.37$\times$ faster when $\alpha~=~0.3$.
Hence, when the degree of ``non-IIDness'' of the data increases (corresponding to decreasing $\alpha$
as explained in Section~\ref{sec:alphaexplanation}), \ours\
performs better. This reduction in the number of rounds not only saves time
but also results in much lower communication costs, as a result of having to
participate in much fewer rounds. In the case of the HAM10000 dataset in Table~\ref{sk:rounds},
the performance benefits of \ours\ is more pronounced. Speedup (fewer rounds) is 1.32-1.52$\times$ for $\alpha~=~0.6$ and 1.56-2.10$\times$ for $\alpha~=~0.3$.
We see a similar trend in the case of FedProx in Tables~\ref{ecg:rounds:prox} and \ref{sk:rounds:prox},
with 1.12-2$\times$ speedup for $\alpha~=~0.6$ and 
1.35-2.14$\times$ speedup for $\alpha~=~0.3$ for MIT-ECG and HAM10000 datasets.

Also, in a world where training jobs are rerun for 1-2 \emph{percentage point}, improvements in accuracy
(in absolute terms), Tables~\ref{mit:accuracy} and \ref{sk:accuracy} for FedYogi illustrate that
\ours\ improves peak model accuracy by $>30$ percentage points for MIT-ECG and $>20$ percentage
points for HAM10000 when $\alpha~=~0.3$ vs. random selection in the case of FedYogi. 
Even for a ``less non-IID'' distribution corresponding to $\alpha~=~0.6$ \ours\ improves accuracy by more than 
12 and 15 percentage points for MIT-ECG and HAM10000, respectively. These benefits
endure when compared to GradClus ($\approx$ 8-30 \% point improvements in accuracy) and TiFL ($\approx$ 6 - 30 \% point improvements in accuracy).
They are lower when compared to OORT, but still significant -- $\approx$ 3-15 and 2-5 percentage points 
corresponding to $\alpha$ of 0.3 and 0.6, respectively. A similar trend 
is seen for FedProx and FedAvg in Tables~\ref{ecg:accuracy:prox}, \ref{sk:accuracy:prox},
\ref{ecg:accuracy:avg} and \ref{sk:accuracy:fedavg}
where accuracy of \ours\ increases by tens of 
percentage points vis-a-vis random selection, TiFL and GradClus.
Unlike in FedYogi and FedProx, where the peak accuracy of \ours\ is higher than that of OORT, peak accuracies of OORT are closer to \ours\ in the case of FedAvg, and
significant in many cases as illustrated in Tables~\ref{ecg:rounds:fedavg}--\ref{sk:accuracy:fedavg}.

Next, we move on to the FEMNIST dataset. This dataset is more IID in its centralized version, and in Table~\ref{femnist:yogi:rounds} we observe that for all the party selection techniques, the target accuracy is reached within the threshold of 200 rounds. The highest accuracy obtained is 86.86 \% for \ours. In Table~\ref{femnist:yogi:rounds}, we can see that \ours~ achieves the target accuracy 1.3x faster than the most comparable OORT technique for $\alpha=0.3$. While for $\alpha=0.6$, OORT reaches the target accuracy almost at the same number of rounds as \ours\ . ~\ours\ is 1.5x - 2.9x faster than Random selection, 1.3x - 2.7x faster than GradCls and 1.4x - 2x faster than TiFL. In Figure~\ref{fig:femnist:all}, we can see that for $\alpha=0.3$, \ours\ performs better than all the other techniques, while for $\alpha=0.6$, OORT performs similar to \ours\ .

For the Fashion MNIST dataset, \ours\ performs better than all the other techniques this can be seen in Figure\ref{fashion:yogi:rounds}. All the techniques attain the target accuracy in the given rounds threshold. \ours\ is 1.2x - 1.5x faster than Random selection, 1.45x - 2x faster than OORT, 1.13x - 1.66x faster than GradCls and 1.73x - 2.2x faster than TiFL. \ours\ attains the highest accuracy 85.14 \%, which is higher than all the other techniques. This is consistent across all the other FL algorithms. 

We also observe that \ours\ achieves the highest model accuracy in all scenarios.

\begin{figure*}[h]
\centering
\subfloat[Accuracy of detecting arrhythmia]{{\includegraphics[width=0.45\textwidth]{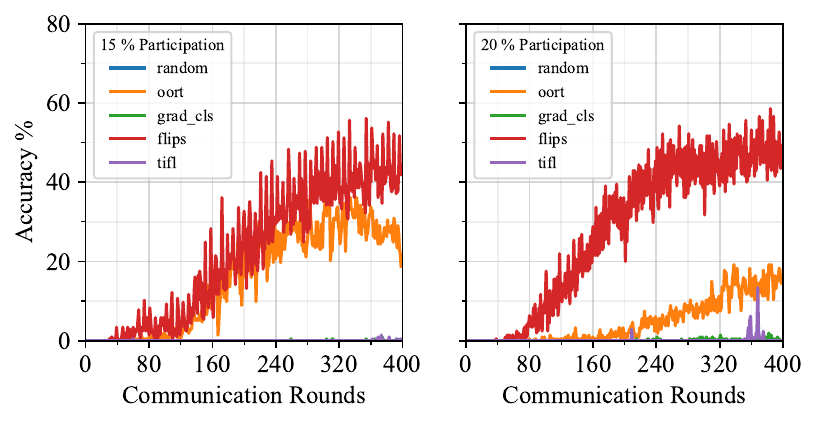}}%
    \label{fig:underrep:ecg}
    }
    \quad
    \subfloat[Accuracy of bcc skin cancer label]{\includegraphics[width=0.45\textwidth] {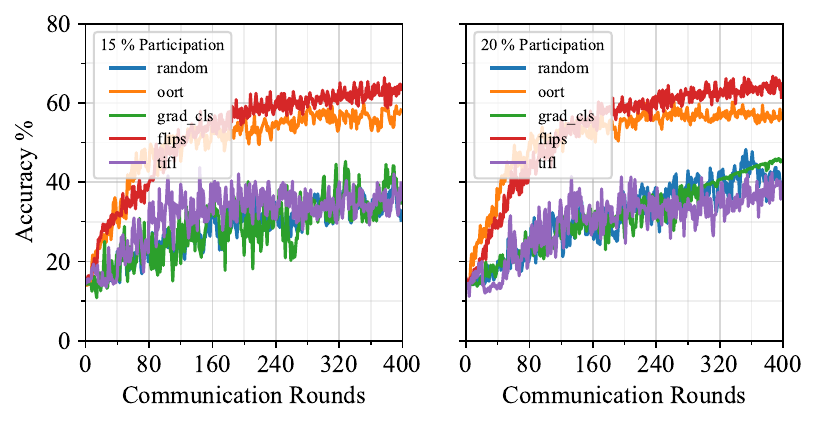}%
    \label{fig:underrep:skmnist}
    }

  \caption{Convergence curves on underrepresented labels for ECG and HAM10000 datasets}
  \label{figure:underrep}
\end{figure*}

This improvement in accuracy can be accredited to the fact that \ours~improves the accuracy of the underrepresented labels. This can be seen in Figure~\ref{figure:underrep}. We can see that \ours~ brings in the most accuracy improvement for the underrepresented labels.

\subsection{Platform Heterogeneity} 

We take the best performing techniques \ours~and OORT and evaluate them under different platform heterogeneity constraints. OORT selects 1.3x the parties in FL at each round to overprovision for straggler parties.

In Table~\ref{mit:rounds} and~\ref{mit:accuracy}, we can observe that under 10 \% Straggler rate OORT is unable to attain the target accuracy even after 400 rounds of training, while \ours~attains the target accuracy 3 out of 4 times while missing the target accuracy by a mere 2.8 \% when $\alpha=0.6$ and party \% = 20. Under the 20 \% straggler rate, the results are similar, OORT cannot reach the target accuracy in all 4 settings, while \ours attains the target accuracy 3 out of 4 times. \ours\ attains the accuracy of 74.66 \%, which is the highest in presence of 10 \% stragglers and 74.20 \% in 20 \% stragglers. This is approximately 4 percentage points less than the 0 \% straggler rate or the ideal scenario. In Table~\ref{mit:rounds} and Table~\ref{mit:accuracy}, we can see that for 10 \% straggler rate, \ours~is faster by 1.3x - 2x than OORT except for the $\alpha=0.6$ and party \% = 20 condition, and attains 11-38 percentage points higher accuracy than OORT. For the 20 \% Straggler rate, \ours~is faster by 1.2x - 2x than OORT except for the $\alpha=0.6$ and party \% = 15 condition, and attains 2 \% - 30 \% higher accuracy than OORT. Figure~\ref{fig:mit:strg:all} also shows that \ours is more robust to stragglers than OORT.

For the HAM10000 (Skin Lesion) Dataset, \ours\ outperforms OORT across all straggler rates. This can be seen in Tables~\ref{sk:rounds} and~\ref{sk:accuracy}. OORT cannot reach the target accuracy in all cases, while \ours~does in all. \ours\ is 1.2x - 1.9x faster than OORT under 10 \% straggler rate, while it is 1.1x - 2.1x faster than OORT under 20 \% straggler rate. \ours\ attains 13 - 17 \% higher accuracy in classification of skin lesions under 10 \% straggler rate and 14 - 26 \% higher accuracy under 20 \% straggler rate. The Figure~\ref{fig:sk:strg:all} depicts the better performance of \ours\ under stragglers via convergence curves.

In the case of the FEMNIST dataset in Table~\ref{femnist:yogi:rounds}, \ours~outperforms OORT and TiFL, when $\alpha=0.6$, the performance is comparable to OORT. For $\alpha=0.3$, \ours~is 1.3x - 1.4x faster than OORT and 1.15x to 1.85x faster than TiFL, while for $\alpha=0.6$, OORT's performance is similar to \ours~. This is because the data distribution is more IID for the 10 \% Straggler rate. For 20 \% \ours\ is 1x - 1.3x faster than OORT and is 1.15 - 1.25x faster than TiFL.   Additionally, \ours\ achieves 1.5 - 3.5 \% higher accuracy than OORT, for a 10 \% straggler rate, while for the 20 \% straggler rate \ours\ attains higher accuracy in all the cases by 0.7 - 2 \% as seen in Figure~\ref{femnist:yogi:rounds}.

For Fashion MNIST we observe that~\ours outperforms OORT and TiFL in all settings. It achieves 3 - 4 \% points improvements in accuracy as seen in Table~\ref{fashion:yogi:accuracy} and is 1.3 - 2.2 faster than OORT and TiFL. These improvements are consistent across FedAvg and FedProx too.



\ours~ is robust against data heterogeneity and platform heterogeneity (presence of stragglers) where data distributions are non-IID. Using intelligent party selection, we improve the terminal accuracy and reduce the communication overheads in FL by using fewer rounds than other techniques. 

To summarize \ours~provides a middleware support system for FL by:
\begin{itemize}[topsep=0pt,partopsep=0pt,parsep=0pt,itemsep=0pt]
    \item Improving the terminal accuracy when the communication overheads (FL rounds) are fixed.
    \item Reducing communication overheads required to attain the desired/target accuracy.
    \item Reducing the time required for FL to attain a target accuracy by lowering the number of FL rounds required.
\end{itemize}

\section{Related Work}

The Intelligent Participant Selection in \ours~deals both with platform and data heterogeneity using a data-driven approach to cluster similar parties and outperforms existing techniques like OORT and GradCls.

\textbf{Data Heterogenenity}: Non-IID datasets in FL introduce client drift in the global model, which hampers its convergence. Many solutions exist to reduce client drift and solve data heterogeneity. ~\cite{pmlr-v119-karimireddy20a, https://doi.org/10.48550/arxiv.2008.03606} propose a client-drift correction update $(m - m_{p})$ between the server model ($m$) and each party's model $m_{p}$ to mitigate client drift. This improves convergence.~\cite{gao2022federated} introduces a penalty and gradient correction term in the local loss function to account for the local drift to bridge the gap between the $m$ and $m_{p}$ for faster convergence in non-IID settings.~\cite{https://doi.org/10.48550/arxiv.2111.04263}  performs more optimizations on the local/party level by introducing dynamic regularization terms to bring the global and local models closer, reducing client drift.

Several studies have examined clustering techniques to personalize models in FL by grouping similar parties in FL based on local model similarity~\cite{Tan_2022}.~\cite{9762352} discuss clustering local model updates using cosine similarity to identify parties with similar data distributions and address data heterogeneity. They create personalized models for each group of parties after a fixed interval, involving re-clustering periodically. IFCA~\cite{NEURIPS2020_e32cc80b} assigns each party a cluster identity based on their data and tailors model parameters for each cluster using gradient descent to improve model convergence for similar parties.

FedLabCluster~\cite{9588126} performs label presence clustering, aggregating models within each cluster to enhance convergence for models specific to each cluster. Another study~\cite{https://doi.org/10.48550/arxiv.2205.15564} investigates sharing encoded data among parties and clustering datasets using K-Means to achieve high clustering accuracy in FL. CMFL~\cite{8885054} dynamically identifies irrelevant local model updates by comparing them to the global model, and discards updates irrelevant to the global model, addressing data heterogeneity in non-IID cases. In a hierarchical federated learning (HFL) setting,~\cite{9546457} improves convergence by selecting parties with the lowest KL divergence between local and edge aggregator's data. FedCBS~\cite{zhang2022fedcbs} computes the Quadratic Class-Imbalance Degree from label distributions to choose parties with more balanced grouped datasets, addressing data heterogeneity. However, none of these studies focus on intelligent participant selection or discuss confidentiality approaches in FL.


\textbf{Platform Heterogeneity}: Solutions like Aergia~\cite{10.1145/3528535.3565238} deal with platform heterogeneity by offloading the training of stragglers to other parties with similar datasets. This reduces the time required for federated learning while maintaining accuracy similar to the baseline techniques.  \ours~leverages the knowledge of similar parties to perform participant selection across parties to make a 2-fold improvement across accuracy and training time, using lesser communication rounds in intermittent FL scenarios.~\cite{10.1007/978-3-031-28996-5_6} solves the straggler issue by using Locality Sensitive Hashing to cluster models/parties and drops duplicate and slow model updates. Further, it requires parties to train locally without knowing whether their models will be used, often wasting local computing resources, which may be undesirable in edge settings. FedLesScan~\cite{https://doi.org/10.48550/arxiv.2211.05739} a clustering framework clusters parties into three groups: rookies, participants and stragglers based on the device variations. Parties are selected from these clusters to mitigate the effect of stragglers to reduce training time and cost.~\ours~ uses a more data-driven approach to pick to compensate for the straggler parties. ~\cite{9887795} uses a mechanism where it selects faster parties in the beginning to attain a target accuracy and then incorporates the slower parties to improve the global model.~\cite{Park2021SageflowRF} deals with platform heterogeneity by using model grouping and weighting based on arrival delay to identify stragglers and entropy-based approaches to mitigate adversaries. ~\cite{10042414} uses gradient coding to introduce redundancy in model training to mitigate the effect of stragglers. FedAT~\cite{10.1145/3458817.3476211} uses a straggler-aware, weighted aggregation heuristic to solve the platform heterogeneity issue. This heuristic assigns higher weights to faster devices, which helps to compensate for the slower devices. FedCS~\cite{10078005} is a communication-efficient federated learning framework inspired by quantized compressed sensing. It compresses gradients at client devices using block sparsification, dimensional reduction, and quantization. Then, it reconstructs gradients at the parameter server and achieves almost identical performance as no compression, while significantly reducing communication overhead.

\section{Towards \ours\ in real deployments}

We implement \ours\ in smartspaces and assisted living for older adults, analyzing real-time data from multiple facilities and individuals. We aim to monitor residents, identify health and safety incidents, and detect changes in daily activities, falls, illnesses, and wandering events. By utilizing federated learning (FL) and \ours, we ensure robust and timely analysis of personal health records and device data while maintaining data privacy.

A specific area of interest is using ECG data from portable devices to detect arrhythmias and heart irregularities. Machine learning models trained on ECG datasets have shown promise in early detection and treatment of cardiac issues. With \ours, we can train heart rhythm and fall models without compromising sensitive personal data, thus enhancing both privacy and model robustness. We have partnered with a senior-care community consisting of approximately 50 facilities, which serves as a trusted entity for storing confidential resident information. These communities act as aggregators and trusted parties, facilitating label distribution in a FLIPS deployment.

Additionally, we focus on the detection and localization of falls in assisted living facilities using data from cameras, acoustic sensors, and wearable tags with accelerometers, gyroscopes, and location-based sensors. Our efforts involve developing robust event detection models to handle device heterogeneity and variations in fall risk~\cite{ALHASSOUN201996}. Privacy concerns are addressed by ensuring secure collection, communication, and storage of remotely obtained data to prevent unauthorized access and misuse.

While \ours\ employs federated and decentralized learning, the clustering aspect is centralized and demonstrated using a centralized aggregator in Section~\ref{sec:flips}. We chose this approach because it aligns with the most commonly used architecture in real-world federated learning systems, such as Google FL~\cite{bonawitz2017practical, bonawitz2019towards}, IBM FL~\cite{ibmfl, ibmflpublic}, FATE~\cite{fate}, and Facebook FL~\cite{huba2022papaya}. The centralized aggregator offers simplicity, statelessness, fault tolerance, and ease of storing FL job data and accounting information in fault-tolerant cloud object stores or key-value stores. In case of aggregator failure, data can be recovered, and aggregation can be resumed from the last round. Communication and aggregator failures are easily recovered by requesting retransmission of lost model updates from the parties, assuming each party securely stores its local model.

\section{Conclusions and Future Work}
In this article, we present FLIPS: Federated Learning using Intelligent Participant Selection, which improves label and participant representation in FL to improve convergence, attain higher accuracy and reduce communication overheads. Our empirical evaluation indicates that \ours~on an average speeds FL algorithms by  1.2$\times$ - 2.9$\times$ and improves terminal accuracy by 5-30 percentage points when compared to the existing techniques.

We are currently exploring the following research directions: (1) personalization in Federated Learning (FL), (2) handling changing data distributions (3) decentralized clustering of label distributions and (4) \ours\ and adversarial robustness. Personalization in FL involves adapting to individual users' or devices' unique requirements and characteristics. Instead of a centralized dataset, we plan to train the model using data from similar parties or devices separately, allowing for personalized models that account for specific patterns and differences in each party's or device's data. Additionally, we plan to address the issue of changing data distributions, which is relevant in IoT settings with streaming data that can introduce shifts in data distributions. \ours~, will capture these changes and train robust models while optimizing resource usage. We anticipate that as datasets and AI-based analytics techniques continue to expand, the ability to handle multiple requirements like privacy, performance, and latency will become crucial, and \ours\ represents a step in that direction.

Decentralized FL architectures rely on secure multi-party computation (SMPC) with or without homomorphic encryption to ensure model update privacy. However, their higher computational requirements make them less popular. To implement \ours\ using SMPC, Algorithm~\ref{alg:fed:flips}, clustering must be computed using an SMPC protocol. Participant selection can be achieved through leader election, with the leader implementing the \ours\ selection protocol and other parties auditing the process. Finally, we
are interested in investigating the interplay between diverse datasets in FL and techniques used for adversarial robustness -- whether
such techniques~\cite{rofl,robust2} exclude valid underrepresented data classes. We also plan to investigate changes needed 
to make label distribution clustering work with adversarially robust FL.




\bibliographystyle{plain}
\bibliography{icdcs-private-clustering}

\end{document}